\pdfoutput=1
\documentclass[11pt]{article}

\usepackage[preprint]{acl}

\usepackage{times}
\usepackage{latexsym}

\usepackage[T1]{fontenc}
\usepackage[utf8]{inputenc}

\usepackage{microtype}

\usepackage{inconsolata}

\usepackage{amsmath,amsfonts,bm}

\def\eqref#1{equation~\ref{#1}}
\def\1{\bm{1}}

\DeclareMathAlphabet{\mathsfit}{\encodingdefault}{\sfdefault}{m}{sl}
\SetMathAlphabet{\mathsfit}{bold}{\encodingdefault}{\sfdefault}{bx}{n}

\newcommand{\E}{\mathbb{E}}

\newcommand{\Var}{\mathrm{Var}}

\newcommand{\Cov}{\mathrm{Cov}}
\DeclareMathOperator*{\argmax}{arg\,max}
\DeclareMathOperator*{\argmin}{arg\,min}

\DeclareMathOperator{\sign}{sign}

\usepackage{xcolor}
\definecolor{thedarkblue}{RGB}{0,0,120} %104} % 180
\definecolor{mydarkblue}{rgb}{0,0.08,0.45} %ICML dark blue
\definecolor{darkblue}{rgb}{0,0.08,180}
\colorlet{TufteRed}{red!80!black}
\definecolor{theblue}{RGB}{0,0,180}
\colorlet{thered}{TufteRed}
      
\usepackage{microtype}
\usepackage{balance}

\usepackage{booktabs}
\usepackage{tabularx}

\usepackage{amsmath,amssymb,amsthm}

\newcommand{\eat}[1]{\ignorespaces}
\usepackage{comment}

\newcommand{\journal}[1]{} % text that is too long for short conf paper, but would be good for journal

\usepackage{tikz}
\usepackage{verbatim}
\usetikzlibrary{arrows}
\usetikzlibrary{shapes,snakes}
\usetikzlibrary{decorations.pathmorphing} % noisy shapes
\usetikzlibrary{fit}					% fitting shapes to coordinates
\usetikzlibrary{backgrounds}	% drawing the background after the foreground

\usepackage{ragged2e}
\usepackage{multirow}
\usepackage{microtype}
\usepackage{balance}
\usepackage{setspace}

\graphicspath{{./}{./graphics/}}
\newcolumntype{H}{>{\setbox0=\hbox\bgroup}c<{\egroup}@{}}

\newcolumntype{R}[1]{>{\RaggedLeft\arraybackslash}} %p{#1}}
\newcolumntype{L}[1]{>{\RaggedRight\arraybackslash}} %p{#1}}

\newcommand{\eg}{\emph{e.g.}}
\newcommand{\ie}{\emph{i.e.}}

\AtBeginEnvironment{pmatrix}{\setlength{\arraycolsep}{2pt}}

\ifdefined\argmax
\else
  \DeclareMathOperator*{\argmax}{argmax}
\fi
\ifdefined\argmin
\else
  \DeclareMathOperator*{\argmin}{argmin}
\fi

\ifdefined\E
\else
  \DeclareMathOperator{\E}{E}
\fi

\DeclareMathOperator{\hugeE}{\mbox{\huge\raise-0.3ex\hbox{E}}}
\DeclareMathOperator{\p}{\mathbb{P}}
\DeclareMathOperator{\hugep}{\mbox{\huge\raise-0.3ex\hbox{$\p$}}}

\ifdefined\Var
\else
  \DeclareMathOperator{\Var}{Var}
\fi

\ifdefined\Cov
\else
  \DeclareMathOperator{\Cov}{Cov}
\fi

\ifdefined\sign
\else
  \DeclareMathOperator{\sign}{sign}
\fi

\usepackage{graphicx}
\usepackage{subfigure}
\usepackage{graphbox}
\usepackage{svg}
\usepackage{nicefrac}

\renewcommand{\cite}[1]{\citep{#1}}

\newcommand{\rev}[1]{\textcolor{blue}{#1}}

\DeclareMathAlphabet{\mathbcal}{OMS}{cmsy}{b}{n}
\usepackage{amsmath}
\usepackage{comment}
\usepackage{bm}
\usepackage{bbm}
\usepackage{amssymb}
\usepackage{enumitem}
\usepackage{tabularx}
\usepackage{multirow}
\usepackage{xcolor}
\usepackage{listings}

\usepackage{wrapfig,lipsum,booktabs}
\usepackage{adjustbox}

\title{FigCaps-HF: A Figure-to-Caption Generative Framework and Benchmark with Human Feedback}

\author{
Ashish Singh$^1$, Ashutosh Singh$^2$, Prateek Agarwal$^1$, Zixuan Huang$^1$, Arpita Singh$^1$,\\ \textbf{Tong Yu$^3$, Sungchul Kim$^3$, Victor Bursztyn$^3$, Nesreen K. Ahmed$^4$, Puneet Mathur$^3$,}\\ \textbf{Erik Learned-Miller$^1$, Franck Dernoncourt$^3$, Ryan A. Rossi$^3$} \\ \\
$^1$University of Massachusetts Amherst, $^2$Northeastern University, $^3$Adobe Research, $^4$Cisco Research \\
\small{\texttt{ashishsinghw@cs.umass.edu, \{dernonco,ryrossi\}@adobe.com.}}
}

\begin{document}
\maketitle
\begin{abstract}
Captions are crucial for understanding scientific visualizations and documents. Existing captioning methods for scientific figures rely on figure-caption pairs extracted from documents for training, many of which fall short with respect to metrics like helpfulness, explainability, and visual-descriptiveness leading to generated captions being misaligned with reader preferences. To address this issue, we introduce \textbf{FigCaps-HF} a new framework for figure-caption generation that can incorporate domain expert feedback in generating captions optimized for reader preferences. Our framework comprises of 1) an automatic method for evaluating the quality of figure-caption pairs, and 2) a novel reinforcement learning with human feedback (RLHF) method to optimize a generative figure-to-caption model for reader preferences. We demonstrate the effectiveness of our simple learning framework by improving performance over standard fine-tuning across different types of models. In particular, when using BLIP as the base model, our RLHF framework achieves a mean gain of 35.7\%, 16.9\%, 9\% and 11.4\% in ROUGE, BLEU, Meteor and CIDEr scores respectively. Finally, we release a large-scale benchmark dataset with human feedback on figure-caption pairs to enable further evaluation and development of RLHF techniques for this problem.
\begin{small}
\\
\textbf{Benchmark:}
\href{https://figshare.com/s/c034fd77bea9475319cb}{Benchmark}
\small
\textbf{Code:}
\href{https://github.com/FigCapsHF/FigCapsHF}{Codebase}\\
\small
\textbf{Documentation:}
\href{https://figcapshf.github.io/}{Documentation}\\
\end{small}

\end{abstract}

%-----------------------------------------------
\section{Introduction}

\begin{figure*}[t]
\centering
\includegraphics[width=0.9\textwidth]{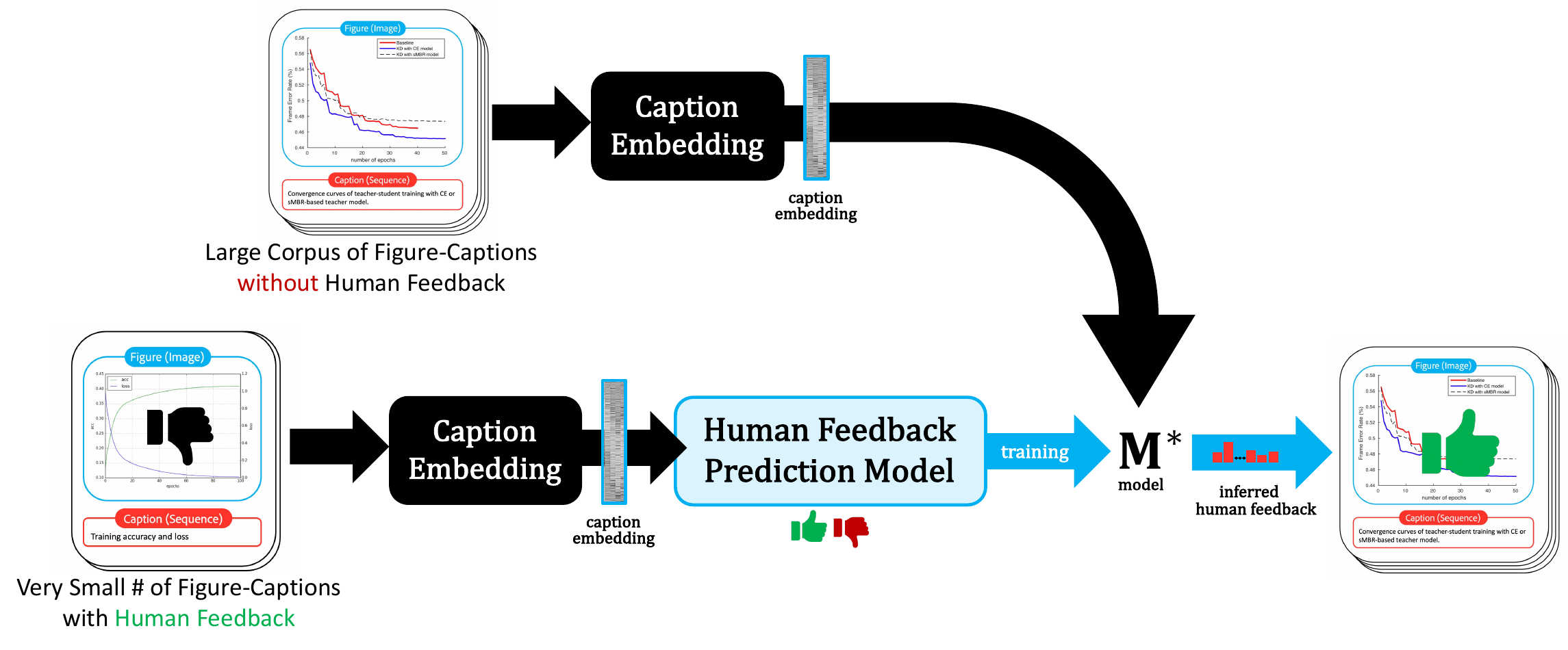}
\caption{%
Our proposed framework for improved figure-captioning using Upside-Down RLHF. The framework utilizes a very small set of reader-feedback annotated figure-caption pairs to learn a calibrated figure-caption scoring model. This model is then used to fine-tune the figure-caption model conditioned on inferred feedback scores.
}%

\label{fig:rlhf-framework}
\end{figure*}

%======================================================================================
For scientific articles, figures (graphs, plots, charts) are integral for conveying key research findings. To understand a given figure and, by extension, the scientific work itself, it becomes crucial that the corresponding captions are informative, i.e., a given caption can represent and complement the figure, situating it in the context of the article. While the importance of figure captions is universally acknowledged, writing a good caption is not trivial. More often than not, many scholarly works contain generic figure captions and lack descriptiveness, thus rendering the figure unhelpful. 

This has motivated extensive research into developing methods that can automatically generate captions for figures to assist researchers in writing better captions.

% Figures such as graphs and plots are crucial in scientific articles for conveying motivation, methodology, and results. Understanding a figure—and by extension, the research—requires informative captions that contextualize the figure within the article. However, writing effective captions is challenging, and many scholarly works contain generic, uninformative captions that limit a figure's utility.

Existing methods treat figure-captioning as a vision-to-language task, where training data is mostly extracted from publically available scientific articles \cite{hsu-etal-2021-scicap-generating}. Many existing datasets, particularly those sourced from platforms like arXiv, contain low-quality captions, which are either uninformative or lack descriptiveness. Such captions can thus result in models with poor generalization and lacking alignment with reader preferences.

% This challenge has motivated research into automated figure-caption generation. Existing methods treat figure-captioning as a vision-to-language task, encoding figure images and metadata to generate captions. These models are trained on figure-caption pairs from scientific articles \cite{hsu-etal-2021-scicap-generating}, but since many captions are poorly written, the resulting models struggle with generalization and fail to align with reader preferences. In \citet{summaries-as-captions-preprint}, over 50\% of captions in arXiv cs.CL papers were deemed unhelpful by domain experts, highlighting the limitations of existing training data.

To address this, we introduce \textbf{FigCaps-HF}, a benchmark and learning framework for improving figure-caption generation by model alignment with reader preferences. Figure~\ref{fig:rlhf-framework} describes our proposed framework designed around two key questions:
% \textbf{(1)} How can we efficiently integrate expert feedback into model training without compromising performance?
\textbf{(1)} How can we integrate expert feedback into model training without additional compute overhead?
\textbf{(2)} How can we scale feedback generation while minimizing human annotation efforts?

For \textbf{(1)}, we employ offline Upside-Down Reinforcement Learning (UDRL) an offline reward-conditioned behavioral cloning method, to align model-generated captions with expert feedback. Once the reward model is trained and generates reward scores, it is no longer needed during figure-caption model training, reducing computational costs while maintaining performance.

For \textbf{(2)}, we develop a caption-rating mechanism guided by reader preference feedback to assess the quality of figure-caption pairs. Using a small, human-annotated dataset with ratings on key factors (e.g., helpfulness, OCR content, takeaway), we train an auxiliary model to predict caption quality scores. This allows us to infer scores for a larger training set, improving scalability. 

Our experimental results demonstrate the effectiveness of our approach. Our trained reward model generalizes well to unseen samples. Evaluations across multiple baseline models show that our reader preference alignment framework outperforms standard supervised fine-tuning, with our best-performing model achieving a 35.7\% increase in BLEU, 16.9\% in ROUGE-L, 9\% in METEOR and 11.4\% in CIDEr scores. Ablation studies further highlight the impact of type and nature of preference feedback on performance.

\medskip\noindent\textbf{Summary of main contributions.} 
\begin{itemize}[leftmargin=*]
\itemsep0em 
\item We introduce an RLHF-based framework for figure-caption generation that uses a small amount of human feedback to train an oracle model, enabling large-scale inference of feedback scores for unseen figure-caption pairs.
\item We develop a method for leveraging limited human feedback to predict feedback scores for new figure-caption pairs, improving model alignment with reader preferences.
\item We release a benchmark dataset to facilitate further research in figure-caption generation with RLHF, fostering advancements in this domain.
\end{itemize}

\section{Background} \label{sec:related-work}

%-------------------------------------------------------------------
\textbf{Figure Caption Generation.}
Initial works in scientific figure captioning focused primarily on model design and feature engineering for caption generation. Works like  \citet{siegel2016figureseer, capunits2021, qian2020formative, chen2019neural, chen2019figure, chen2020figure, hsu-etal-2021-scicap-generating} followed a standard pipeline of utilizing a CNN based vision-encoder to encode figure-features followed by a LSTM/RNN based text-decoder to generate captions. For model training \citet{chen2019neural, chen2019figure, chen2020figure} created and used synthetic figure-caption pairs while in \cite{siegel2016figureseer, hsu-etal-2021-scicap-generating} figure-caption pairs were extracted from publicly available scientific works. With recent advancements in multimodal learning, the standard pipeline has shifted to utilizing pre-trained transformer based vision-language models for either zero-shot inference or supervised fine-tuning on specific domains for image-to-text generation. Recent works like \cite{roberts2024scifibench} have focused on benchmarking large multimodal models (LMMs) for figure-caption generation under zero-shot and fine-tuning settings. In contrast, our work is focused on improving model alignment with respect to reader preference in a simple and scalable manner. Our proposed framework is thus model agnostic and applicable to any LMM.
\\
\textbf{Figure Question Answering.}
A closely related task is Figure Question Answering, which formulates the more general problem of figure understanding as a visual-question answering task. There has been a variety of works in this space towards modeling~\cite{siegel2016figureseer,kahou2017figureqa,li2022multi,singh2020stl,zou2020affinity, kafle2018dvqa, kafle2020answering} as well as creating curated datasets including DVQA~\cite{kafle2018dvqa}, FigureQA~\cite{kahou2017figureqa}, PlotQA~\cite{methani2020plotqa}, Leaf-QA~\cite{chaudhry2020leaf}, and ChartQA~\cite{masry2022chartqa}. In contrast, the proposed framework addresses figure caption generation and does not focus on figure question answering.

\textbf{Learning with Human Feedback} Aligning model predictions with human preference has been shown to improve task performance in various areas, including natural language processing tasks like language model pretraining \cite{korbak2023pretraining}, machine translation \cite{bahdanau2016actor, kreutzer2018can}, text summarization \cite{stiennon2020learning}, unlearning undesirable behaviors from language models \cite{lu2022quark}, computer vision tasks like text-to-image generation \cite{lee2023aligning, zhang2023hive} and reinforcement learning tasks like training RL agents \cite{macglashan2017interactive, ibarz2018reward, lee2021pebble}. In contrast to prior works, we aim at improving figure caption generation by optimizing model learning to align with domain expert feedback. However, unlike previous work that leverages on-policy RL \cite{schulman2017proximal} algorithm to maximize the reward-weighted likelihood, our framework utilizes reward-conditioned behavioral cloning \cite{emmons2021rvs}, an offline variant of upside-down RL method \cite{srivastava2019training} to optimize model learning for reader preference. This provides a simpler and more controllable framework for human preference alignment. Furthermore, our feedback scheme allows for incorporating multiple feedback at different granularity as reward signal during the model optimization step, thus improving model learning.

\section{Framework} \label{sec:approach}

% \rev{%}
% In this section, we introduce our proposed framework for learning with expert feedback (Figure \ref{fig:rlhf-framework}).
% }%
% We first describe a standard figure-captioning pipeline (Sec. \ref{prelim}). Next, we provide details of designing and training a generalizable human-feedback prediction model (Sec. \ref{human-feedback-model}). Finally, we discuss our feedback-aligned model training strategy instantiated as a simple RLHF framework (Sec. \ref{rlhf}).

In this section, we present our framework for learning with expert feedback (Figure \ref{fig:rlhf-framework}). First, we describe a standard figure-captioning pipeline (Sec. \ref{prelim}), then outline the design and training of a generalizable human-feedback prediction model (Sec. \ref{human-feedback-model}), and conclude with our feedback-aligned model training strategy using a simple RLHF framework (Sec. \ref{rlhf}).

\subsection{Preliminaries}\label{prelim}

Given the dataset $D_w$, we can then define a model $f_\theta$, that takes in information corresponding to the figure and outputs a sequence of text as output.

Model $f_\theta$ consists of a vision encoder module to get image-based encoding and a language encoder-decoder module to encode and generate corresponding text. The weights $\theta$ can either be randomly initialized, or initialized by large-scale pretrained model weights. Furthermore, the model weights corresponding to the vision encoder and text encoder-decoder models can either be initialized with separate weights or jointly trained model weights. After initialization, model $f_\theta$ can then be trained for the task of caption generation.

Generally, for training such a model, Language Modeling (LM) loss is used as a standard training objective. 
Let $\{ I_i, T_i \} \in D$ be the input to the model $f_\theta$, where $ I_i \in \mathbb{R}^n$ is the input figure, and $T_i$ is the corresponding text sequence. Additionally, $T_i$ is represented as sequence of $K_j$ tokens from a fixed vocabulary $\mathcal{V}$: $T_i = (T_{i,1},...T_{i,K_j})$, where $K_j = |T_i|$. 
Then the training objective is defined as:
\begin{equation}\label{obj:lm}
    \mathcal{L}_\text{LM} = \frac{1}{K_j +1}\sum\limits_{j=0}^{K_j + 1} H(T_{i,j} | I_i, (T_{i,0}, ...,T_{i,j-1})) ,
\end{equation}
where H denotes the cross-entropy loss and $(T_{i,0}, ...,T_{i,j-1})$ represents all the tokens in the caption prior to $T_{i,j}$.

\subsection{Human Feedback Prediction Model} \label{human-feedback-model}
To improve figure-caption generation, we propose to incorporate domain expert feedback into our optimization step. To generate feedback for figure-caption pairs, we thus propose to learn a feedback prediction model to score individual datasample based on different metrics representing reader preferences. Our objective is to learn a model that can predict human feedback scores for unseen captions accurately given small set of training samples.

To this end, we first label a small control set $D_h$ consisting of $M$ figure caption pairs $\{ I_w, T_w \}$ with domain experts ratings. Here we assume that $M \ll N$, i.e. the size of the control set is significantly less than the original noisy dataset. We can now train a model on $D_h$ to predict the human expert ratings for the original dataset $D_w$. 
Specifically, given human feedback dataset $D_h$ containing figure-caption pairs $\{ I_h, T_h \} \in D_h$ and $k$ human expert evaluation metrics for each datasample $y_i \in \{y_0, y_1, ...y_k\}$, we want to train $k$ models $R(x_i, \theta)_k$ to predict the $k$ scores, respectively. Here the output of a model $R(x_i, \theta)_k(T_h)$ is a scalar quantity denoting a specific metric score for the given input caption. Thus we formulate the scoring problem as a regression task.
Specifically, we can define our human-feedback prediction model as follows:
\begin{equation}\label{obj:reward}
    R(x_i, \theta)_k(T_h) = g(l(\theta_l, x_i), \theta_g),  
\end{equation}
where, $R(x_i, \theta): \mathbb{R}^N \rightarrow \mathbb{R}$,  $l(x_i, \theta_l): \mathbb{R}^N \rightarrow \mathbb{R}^D$ and $g(u_i, \theta_g): \mathbb{R}^D \rightarrow \mathbb{R} $. In the above, $l(., \theta_l)$ is an embedding function that takes in input data $x_i \in \mathbb{R}^N$ and generates corresponding representation $u_i \in \mathbb{R}^D $, and $g(., \theta_l)$ is a regression function to generate the scores respectively. We only train the regression function while keeping the weights of the embedding function fixed. For training the regression function, we use mean-squared error loss, written as: $  \mathcal{L}_\text{R} = \frac{1}{D_{h}}\sum_{i=1}^{D_h}(\hat{y_i} -y_i)^2,$ where $\hat{y_i}$ is the predicted score while $y_i$ is the ground-truth evaluation score. 
After training the human-feedback prediction models, we compute scores for all the samples in the training dataset $D_w$ to construct our new set, which will be used for training the figure-caption model.

\subsection{Reinforcement Learning with Human Feedback}\label{rlhf}

We use the human-feedback prediction model as a reward model to train an image-to-text model for generating higher-quality captions, framing the problem as a reinforcement learning task. Given a dataset $D_w$ with figure-caption pairs $\{ I_w, T_w \}$, we treat figures $I_w$ as states, captions $T_w$ as actions, and predicted metric scores $R(T_w)$ as rewards. Our goal is to train an image-to-text model $f(\theta)$ that maps states to actions while maximizing rewards, ensuring that captions align with human judgment.

We adopt offline UDRL for its computational efficiency and robustness~\cite{emmons2021rvs}. Here, the policy $\pi_\theta$ maps states ($S_t$) to actions ($a_t$) given rewards ($r_t$), formulating learning as a supervised problem. We sample triplets $\{S_t, a_t, r_t\}$ to construct a dataset and train $\pi_\theta$ using:
\begin{equation}
    \max_{\theta}\sum_{t \in D} \mathbb{E} [ \log \pi_\theta (a_t | S_t, r_t)]
\end{equation}

Following this UDRL framework, we define our figure-to-caption model $f(\theta)$ as the policy $\pi_\theta$. For each caption $T_i$, we compute a reward score and binarize it into control tokens: \texttt{<|good|>} if $R(I_i, T_i) \geq t$, otherwise \texttt{<|bad|>}, where $t$ is a hyperparameter. Given this feedback, we fine-tune $f_\theta$ using:
\begin{equation}
    \mathcal{L}_\text{HF} = \frac{1}{K_j +1}\sum\limits_{j=0}^{K_j + 1} H(T_{i,j} | I_i, (c_i,T_{i,0}, ...,T_{i,j-1}))
\end{equation}
where $c_i$ is the control token derived from $R$.

\section{FigCaps-HF: Figure-Captioning with Human Feedback Benchmark}

% train, validate, test
% 106,834 	13,354 	13,355
% train=106,834 	validate=13,354 	test=13,355
\begin{table*}[t]
\centering
\renewcommand{\arraystretch}{1.15} 
\small
% \footnotesize
% \scriptsize
% \begin{tabularx}{1.0\linewidth}{l @{}c HH HH cc cc cccc}
\begin{tabularx}{0.9\linewidth}{l @{}c l c cc Hc c H HH HH}
\toprule
% & 
% % \multirow{2}{*}{\sc \# Trainable} 
% &
% \multicolumn{2}{c}{ROUGE-1} & 
% \multicolumn{2}{c}{\sc ROUGE-2} &
% % &&&&
% \multicolumn{2}{c}{\sc ROUGE-L} & \\
% \cmidrule(r){3-4} \cmidrule(r){5-6} \cmidrule(r){7-8}

% INFERRED HUMAN FEEDBACK
% | Category  | Median    | Mean      | Standard Deviation | Minimum   | 25th Percentile | 75th Percentile | Maximum   |
% |-----------|-----------|-----------|--------------------|-----------|----------------|----------------|-----------|
% | Helpfulness | 2.887949 | 2.891968 | 1.067641           | -1.271212 | 2.167202       | 3.610363       | 7.605054 |
% | Takeaway  | 1.948700 | 2.058931 | 1.028708           | -1.061640 | 1.330719       | 2.660354       | 8.144583 |
% | Visual    | 1.913612 | 2.024573 | 1.010477           | -1.233484 | 1.307569       | 2.625947       | 7.971579 |
% | OCR       | 3.882775 | 3.840144 | 0.828866           | 0.017950  | 3.319476       | 4.406052       | 7.505501 |

% & \textbf{\# Fig-Caption} & \\
% & \textbf{Pairs} & 
& \textbf{\# Fig-Caption Pairs} &
\textbf{Human Feedback}  & \textbf{Median}    & \textbf{Mean}      & \textbf{Std} & \textbf{Min}   & \textbf{Q1} & \textbf{Q3} & \textbf{Max}   &\\
\midrule

\multirow{3}{*}{\textsc{Actual}} % HF (400)}}
& \multirow{4}{*}{\textbf{438}}
% ACTUAL HUMAN FEEDBACK
	   % Median	Mean	Std	Min	25%	50%	75%	Max
&  \textbf{Helpfulness}	 & 3	 & 3.01	 & 1.19	 & 1	 & 2	 & 3		 & 5\\
 \multirow{3}{*}{\textsc{Human Feedback}}
& & \textbf{Takeaway}	 & 	2 & 	2.16 & 	1.22 & 	1	 & 1	 & 2 & 		5\\
& & \textbf{Visual}	&	2	 & 2.11	 & 1.08	 & 1 & 	1	 & 2	 &  	5\\
& & \textbf{OCR}	 & 	4	 & 3.83 & 	0.80 & 	1	 & 4 & 	4	 &  	5\\

\midrule
% train=106,834 	validate=13,354 	test=13,355
\multirow{3}{*}{\textsc{Predicted}} % (100K+)}}
& \multirow{4}{*}{\textbf{106,834}} % (100K+)}}
& \textbf{Helpfulness} & 2.89 & 2.89 & 1.07           & -1.271212 & 2.17       & 3.61       & 7.605054 &\\
\multirow{3}{*}{\textsc{Human Feedback}}
&& \textbf{Takeaway}  & 1.95 & 2.06 & 1.03           & -1.061640 & 1.33       & 2.66       & 8.144583 &\\
&& \textbf{Visual}    & 1.91 & 2.02 & 1.01           & -1.233484 & 1.31       & 2.63       & 7.971579 &\\
&& \textbf{OCR}       & 3.88 & 3.84 & 0.83           & 0.017950  & 3.32      & 4.41       & 7.505501 &\\

\bottomrule
\end{tabularx}
\caption{%
Summary of our benchmark dataset for figure-caption generative models with RLHF.
}
\label{table:benchmark-data-stats}
\end{table*}

% \begin{table*}
% % \begin{adjustbox}{width=\columnwidth}
% \begin{tabular}{lll}
% \hline
% \textbf{Filtering method} & \textbf{CER \%} & \textbf{WER \%}\\
% \hline
% Original Model & 24.22 & 35.76 \\
% Prediction Uncertainty & 19.43 & 26.63 \\
% Phrase Log-Likelihood & 21.4 & 31.5 \\
% Union & 20.25 & 30.43 \\
% Intersection & 16.47 & 25.64 \\
% \hline
% \end{tabular}
% % \end{adjustbox}
% \caption{Evaluation of OCR model after fine-tuning on that target dataset.}
% \label{t2}
% \end{table*}

We propose a new benchmark for figure-captioning with feedback. Our benchmark consists of 106,834 figure-caption pairs \cite{hsu-etal-2021-scicap-generating} with feedback scores.
Our dataset contains feedback based on different measures to evaluate quality of the author written captions for the corresponding figure. For each figure-caption pair, we evaluate the data sample based on four quality measures: \textbf{(1) Helpfulness}, \textbf{(2) Takeaway}, \textbf{(3) Visual-descriptiveness (visual)} and \textbf{(4) Image-text (OCR)} \cite{summaries-as-captions-preprint}. Each quality metric is selected to measure the ability of the readers to comprehend and draw inferences based on the provided figure and the corresponding caption. 

We compute the feedback scores for each data sample in a scalable manner by first annotating a small subset with domain-expert feedback and then predicting score for the entire dataset using the human-feedback model described in Sec. \ref{human-feedback-model}. Specifically, we select 438 randomly sampled figure-caption pairs, each annotated by domain experts \cite{summaries-as-captions-preprint}. Each pair has been evaluated on 5-point Likert scale for each of the above mentioned quality metric. Using this labeled subset, we train a human-feedback prediction model to generate scores for the remainder of the dataset. Unlike the subset, we keep the scores for the entire dataset as a continuous value. This allows the users of the benchmark to accordingly decide their own scheme for labeling each figure-caption pair based on different thresholding criteria, thus providing flexibility for fine-grained feedback. 

Table~\ref{table:benchmark-data-stats} presents an overview of the statistics related to the actual and predicted human feedback for the captioning of the scientific figures. We see that the predicted human feedback values in our study show a diverse range, as indicated by the small standard deviation of $1\pm0.2$ and a consistent mean value across all ratings. Additionally, the alignment of the median predicted scores with the actual human feedback values indicates that the model's performance is not skewed towards any particular rating but provides an accurate assessment across the range of ratings. This suggests that the human-feedback prediction model used to infer the scores is generalizable and can accurately assess the quality of captions across various ratings. Furthermore, the proposed model provides reliable scores for captions that fall outside the typical range of scores. 

We provide more details in the section \ref{sec:data} in the Appendix.

%======================================================================================

\begin{table*}
\centering
\renewcommand{\arraystretch}{1.15} 
\small
% \footnotesize
% \scriptsize
% \begin{tabularx}{1.0\linewidth}{l @{}c HH HH cc cc cccc}
\begin{tabularx}{0.95\linewidth}{l l c HH HH Hc c c H ccc}
\toprule
% & 
% % \multirow{2}{*}{\sc \# Trainable} 
% &
% \multicolumn{2}{c}{ROUGE-1} & 
% \multicolumn{2}{c}{\sc ROUGE-2} &
% % &&&&
% \multicolumn{2}{c}{\sc ROUGE-L} & \\
% \cmidrule(r){3-4} \cmidrule(r){5-6} \cmidrule(r){7-8}
&
\textsc{Model} & 
% \textsc{Parameters} & 
\textbf{\#Params} & 
R@1 & \sc Norm. & %R@1 & 
R@2 & \sc Norm. & %R@2 & 
R@L & 
% \sc Norm. & %R@L &
\textbf{ROUGE-L} &
\textbf{BLEU} & 
\textbf{CIDEr} & 
\textbf{SPICE} &
\textbf{METEOR} &
\\ \midrule

\multirow{1}{*}{\textsc{ocr-only}}
& \textbf{Pegasus}
% ~\cite{zhang2020pegasus} 
& 0.27B & 0 & 0 & 0 & 0 & 0 & 0.026 & 4.78e-4 & 0.134 & 0.0 & 0.042 \\ 
% & \textbf{Pegasus TODO (pretrained)}~\cite{zhang2020pegasus} & 0.27B & 0 & 0 & 0 & 0 & 0 & 0.026 & 7.8e-5 & 0.01 & 0.0 & 0.042 \\ 
% & \textbf{Pegasus (pretrained)}~\cite{zhang2020pegasus} & 0.27B & 0 & 0 & 0 & 0 & 0 & 0.435 & 0.271 & 1.741 & 0 & 0.413 \\ 

\midrule
\multirow{4}{*}{\textsc{Figure-Only}}
& \textbf{TrOCR}
% ~\cite{li2021trocr} 
& 0.23B & 0 & 0 & 0 & 0 & 0 & 0.025 & <0.001 & 0.016 & 0.0 & 0.018 \\ 

& \textbf{BEiT+GPT2}
% ~\cite{} 
& 0.24B & 0 & 0 & 0 & 0 & 0 &  0.142 & 0.005 & 0.372  & 0 & 0.124 \\ 
% & \textbf{BEiT+GPT2 with BLIP Filter dataset (50 percentile)} & 0.24B & 0 & 0 & 0 & 0 & 0 &  0.127 & 0.002 & 0.000  & 0 & 0.111 \\ 
% & \textbf{BEiT+GPT2 with MCSE Filtered dataset (50\% @ 5 epochs)} & 0.24B & 0 & 0 & 0 & 0 & 0 &  0.132 & 0.003 & 0.054  & 0.0 & 0.121 \\ 
% & \textbf{BEiT+GPT2 with MCSE Filtered dataset (50\% @ 10 epochs)} & 0.24B & 0 & 0 & 0 & 0 & 0 &  0.135 & 0.016 & 0.054  & 0.0 & 0.125 \\ 
& \textbf{ViT
%\cite{dosovitskiy2020vit} 
+ RoBERTA
% \cite{liu2019roberta}
} & 0.23B & 0 & 0. & 0 & 0 & 0 & 0.140 & 0.012 & 0.380 & 0 & 0.121\\ 
% & \textbf{ViT\cite{dosovitskiy2020vit} + GPT2} & 0.24B & 0 & 0 & 0 & 0 & 0 & 0.142 & \Red{0.006} & 0.061 & 0 & 0.126 \\ 
& \textbf{ViT
% \cite{dosovitskiy2020vit} 
+ GPT2} & 0.24B & 0 & 0 & 0 & 0 & 0 & 0.142 & 0.018 & 0.427 & 0 & 0.126 \\
% & \textbf{ViT\cite{dosovitskiy2020vit} + GPT2(beam)} & 0.24B & 0 & 0 & 0 & 0 & 0 & 0.140 & 0.019 & 0.061 & 0 & 0.125 \\

\midrule
\multirow{5}{*}{\textsc{Figure-Caption}}
% https://huggingface.co/vqascore/promptcap-coco-vqa (check if another version exists)
% & \textbf{PromptCap(0-shot)}~\cite{hu2022promptcap} & 0.47B & 0 & 0 & 0 & 0 & 0 & 0.130 & 0.004 & 0.121 & 0 & 0.082\\ 
& \textbf{PromptCap}
%~\cite{hu2022promptcap} 
& 0.47B & 0 & 0 & 0 & 0 & 0 & 0.130 & 0.009 & 0.269 & 0 & 0.082\\

% https://huggingface.co/dhansmair/flamingo-mini
& \textbf{Flamingo}
%~\cite{alayrac2022flamingo} 
& 1.14B & 0 & 0 & 0 & 0 & 0 & 0.087 & 0.001 & 0.243 & 0 & 0.046 \\ 
% & \textbf{Flamingo-mini}~\cite{alayrac2022flamingo} & 1.14B & 0 & 0 & 0 & 0 & 0 & 0.0865 & 0.0002 & 0.0152 & 0 & 0.0459\\ 
% % https://huggingface.co/allenai/scibert_scivocab_uncased
% & \textbf{SciBERT}~\cite{zhang2021vinvl} & 0 & 0 & 0 & 0 & 0 & 0 & 0 & 0 & 0 & 0 & 0\\ 

% & \textbf{GIT~\cite{wang2022git}} & 0.17B & 0 & 0 & 0 & 0 & 0 & 0.165 & 0.021 & 4.8e-7 & 0 & 0.142 \\
& \textbf{GIT}
%~\cite{wang2022git} 
& 0.17B & 0 & 0 & 0 & 0 & 0 & 0.119 & {0.002} & 0.219 & 0 & 0.091 \\
% & \textbf{GIT (fine-tuned)~\cite{wang2022git}} & 0.17B & 0 & 0 & 0 & 0 & 0 & 0.119 & 0.0018 & 4.8e-7 & 0 & 0.0906 \\
% & \textbf{GIT (pretrained)\cite{wang2022git}} & 0.17B & 0 & 0 & 0 & 0 & 0 & 0.106 & 0.001 & 0.0179 & 0 & 0.057 \\ 

& \textbf{BLIP}
%~\cite{li2022blip} 
& 0.25B & 0 & 0 & 0 & 0 & 0 & 0.130 & 0.014 & 0.438 & 0 & 0.132 \\ 
% & \textbf{BLIP \cite{li2022blip}} & 0.25B & 0 & 0 & 0 & 0 & 0 & 0.169 & 0.027 & 0.17 & 0 & 0.139 \\ 
% & \textbf{CLIPCap} & 0.15B & 0 & 0 & 0 & 0 & 0 & 0.078 & 0.004 & 0 & 0 & 0.095 \\ 
& \textbf{CLIPCap}%~\cite{mokady2021clipcap} 
& 0.15B & 0 & 0 & 0 & 0 & 0 & 0.103 & 0.012 &  0.284 & 0 & 0.131 \\ 

\midrule
% \multirow{3}{*}{\textsc{rlhf-prepend}}
\multirow{2}{*}{\textsc{rlhf}}
% \multirow{4}{*}{\textsc{Feedback-Prompt (Prepend)}}
% PREPENDED & 'good' prompt during inference.
% & \textbf{Ours-BLIP-RLHF} & 0.25B & 0 & 0 & 0 & 0 & 0 & 0.1520 & 0.0186 & TODO &  0 & 0.1450 \\ 
& \textbf{Ours-BLIP-RLHF} & 0.25B & 0 & 0 & 0 & 0 & 0 & \textbf{0.152} & 0.019 &  \textbf{0.552} &  0 & \textbf{0.145} \\ 
% & \textbf{Ours-CLIPCap-RLHF} & 0.15B & 0 & 0 & 0 & 0 & 0 & 0.104 & 0.006 & TODO &  0 & 0.115 \\ 
% & \textbf{Ours-CLIPCap-RLHF} & 0.15B & 0 & 0 & 0 & 0 & 0 & 0.098 & 0.010 & TODO &  0 & 0.120\\ 
% & \textbf{Ours-CLIPCap-RLHF} & 0.15B & 0 & 0 & 0 & 0 & 0 & 0.104 & 0.0057 & TODO &  0 & 0.115 \\ 
% & \textbf{Ours-GIT-RLHF (MLP)} & 0.17B & 0 & 0 & 0 & 0 & 0 & 0.157 & 0.023 & TODO & 0 & 0.143 \\ 
% & \textbf{Ours-ViT+GPT2-RLHF} & 0.24B & 0 & 0 & 0 & 0 & 0 & \textbf{0.155} & \Red{0.007} & TODO &  0 & 0.133 \\ 
& \textbf{Ours-ViT+GPT2-RLHF} & 0.24B & 0 & 0 & 0 & 0 & 0 & 0.138 & \textbf{0.020} & 0.489 &  0 & 0.126 \\ 
% & \textbf{Ours-ViT+GPT2-RLHF(beam)} & 0.24B & 0 & 0 & 0 & 0 & 0 & 0.138 & 0.019 & TODO &  0 & 0.126 \\ 

% & \textbf{Ours-BEiT+GPT2-RLHF} & 0.24B & 0 & 0 & 0 & 0 & 0 & 0.135 & 0.017 & TODO &  0 & 0.122 \\ 

% & \textbf{Ours-BLIP-RLHF (MLPRegressor Scoring) } & 0.25B & 0 & 0 & 0 & 0 & 0 & 0.212 & 0.034 & TODO &  0 & 0.18 \\ 
% & \textbf{Ours-VIT-GPT2-RLHF (MLPRegressor Scoring) } & 0.25B & 0 & 0 & 0 & 0 & 0 & 0.151 & 0.007 & TODO &  0 & 0.14 \\ 
% OCR-caption contrastive loss + language fine-tuned (same as BLIP+OCR), inference uses only figure
% & \textbf{Ours-OCR-Caption-CLoss} & 0.27B & 0 & 0 & 0 & 0 & 0 & 0.10794 & 0.000855 & TODO & 0 &  0.09758 \\ 
% OCR-caption contrastive loss + language fine-tuned (same as BLIP+OCR), inference uses only figure+OCR
% & \textbf{Ours-OCR-Caption-CLoss} & 0 & 0 & 0 & 0 & 0 & 0 & 0 & 0 & 0 & 0 & 0 \\ 
% & \textbf{Ours} & 0 & 0 & 0 & 0 & 0 & 0 & 0 & 0 & 0 & 0 & 0 \\ 
\bottomrule
\end{tabularx}
\caption{
Comparison with state-of-the-art methods. 
% \todo{add more context, keep it short}
% \todo{Please make sure all methods are listed correctly, caption-only, figure-only, figure-caption}
%\ryan{I think these numbers need to be multiplied by 100 (See BLIP and other papers)?}
For all the metrics, higher values are better ($\uparrow$).
% \ryan{%
% BLEU=BLEU@4 (add somewhere in paper to clarify)
% }%
}
\label{table:results-comparison}
\end{table*}

% =================================================================================
\begin{figure*}[t]
\centering

\includegraphics[width=0.95\linewidth]{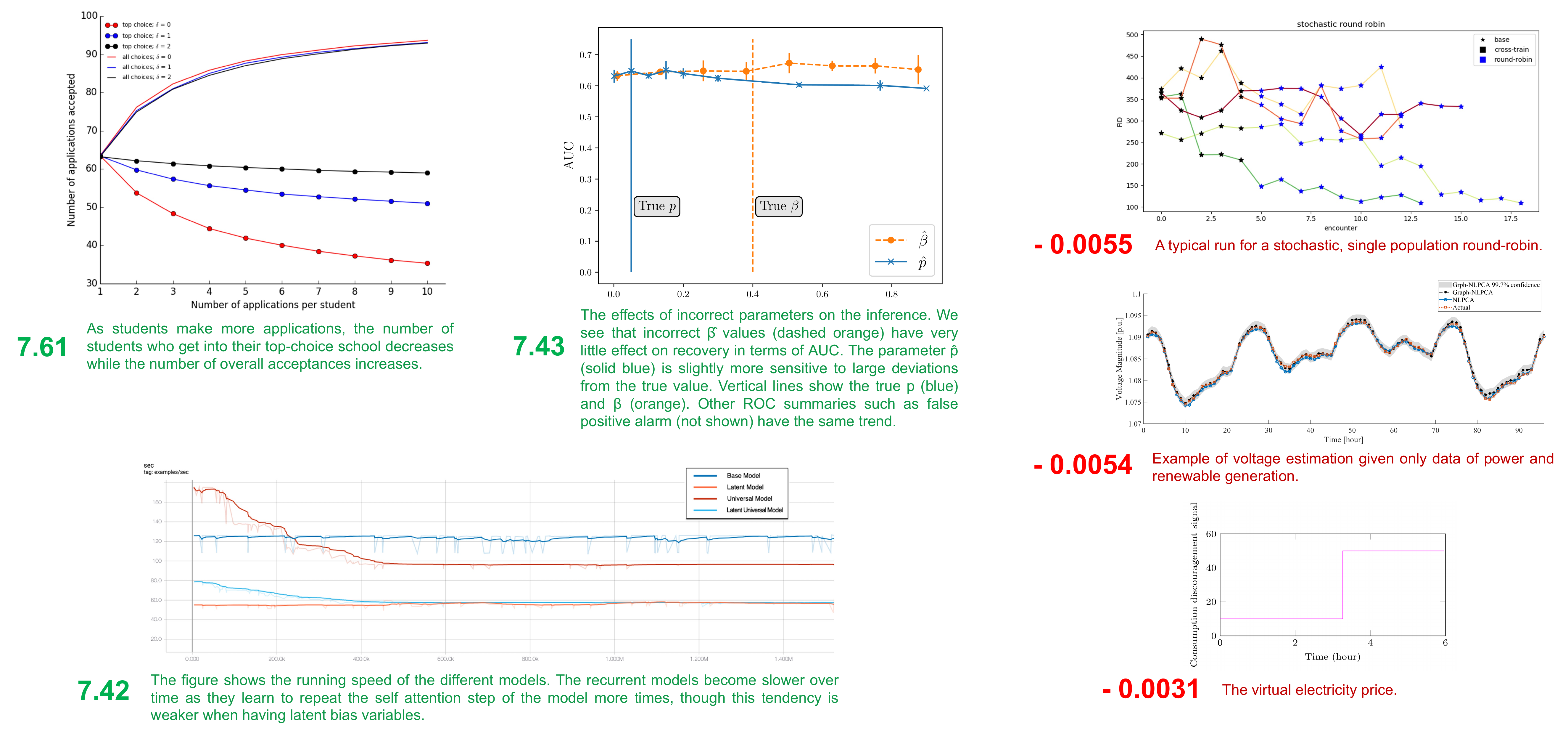}

\caption{Results of our Human Feedback Prediction Model. Here we show the three figure-caption pairs with the highest (left; green) and smallest (right; red) ``helpfulness'' human feedback score from our trained HF model.
Notably, the figure-caption pairs rated highly by our human-feedback predictive model are better as they mention specific takeaways, figure text and visual details. 
In contrast, the figure-caption pairs with lowest scores by our predictive model are those that are extremely vague and uninformative.}
\label{fig:HF-predictive-model-top-and-worst-figs}

\end{figure*}
%------------------------------------------------------------------------------------
\begin{figure*}[ht]
\centering

\includegraphics[width=1.0\linewidth]{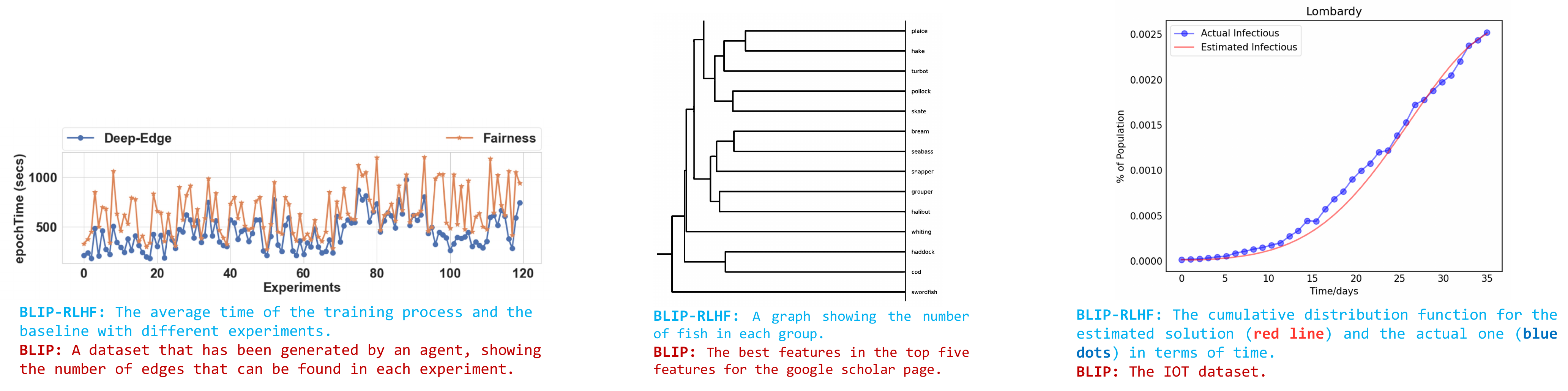}
\vspace{-5mm}
\caption{Generated captions from our RLHF framework using BLIP as the base model (in Blue) compared to BLIP without RLHF (in Red). Fine-tuning BLIP with human-feedback predictions significantly improve the caption quality with respect to descriptiveness while maintaining conciseness.}  

\label{fig:RLHF-vs-base}
% \vspace{-2mm}
\end{figure*}
%------------------------------------------------------------------------------------
\begin{table}[t!]
\centering
\renewcommand{\arraystretch}{1} 
\setlength{\tabcolsep}{3pt} 
\small
% \footnotesize
% \scriptsize
% \begin{tabularx}{1.0\linewidth}{l @{}c HH HH cc cc cccc}
\begin{tabularx}{\linewidth}{l l H HH HH Hc cH Hccc}
\toprule
% & 
% % \multirow{2}{*}{\sc \# Trainable} 
% &
% \multicolumn{2}{c}{ROUGE-1} & 
% \multicolumn{2}{c}{\sc ROUGE-2} &
% % &&&&
% \multicolumn{2}{c}{\sc ROUGE-L} & \\
% \cmidrule(r){3-4} \cmidrule(r){5-6} \cmidrule(r){7-8}
&
\textsc{Feedback} & 
% \textsc{Parameters} & 
\textbf{\#Params} & 
R@1 & \sc Norm. & %R@1 & 
R@2 & \sc Norm. & %R@2 & 
R@L & 
% \sc Norm. & %R@L &
\textbf{ROUGE-L} &
\textbf{BLEU} & 
\textbf{CIDEr} & 
\textbf{SPICE} &
\textbf{METEOR} &
\\ \midrule

% & \textbf{BLIP with Multi-labeled feedback(10 epochs multi label)} 
% & \textbf{BLIP with Multi-labeled feedback(10 epochs multi label)} 
& \textbf{Binary} 
& 0.25B & 0 & 0 & 0 & 0 & 0 & 0.152 & 0.019 & TODO &  0 & 0.145 \\ 

& \textbf{Multi-label}
& 0.25B & 0 & 0 & 0 & 0 & 0 &   0.153 &  \textbf{0.022} & 0.000  & 0 & \textbf{0.151} \\ 
% & \textbf{BLIP with Multi-labeled feedback (5 epoch binary +5 epochs multi label)} 
& \textbf{Binary + Multi-label}
% & \textbf{BLIP (5 epoch binary +5 epochs multi label) \ryan{need better name}}
% , which the name can be described in text}} 
& 0.25B & 0 & 0 & 0 & 0 & 0 &  \textbf{0.156} &  0.019 & 0.0  & 0.0 & 0.148 \\ 
% & 0.25B & 0 & 0 & 0 & 0 & 0 &  \textbf{0.156} &  0.0191 & 0.0  & 0.0 & 0.1484 \\ 
% & \textbf{BLIP with Multi-labeled feedback (bad/neutral/good)} & 0.25B & 0 & 0 & 0 & 0 & 0 &  0.151 &  0.01747 & 0.0  & 0.0 & 0.1438 \\ 
% & \textbf{Ours-BLIP-RLHF} 

\bottomrule
\end{tabularx}
\vspace{-1mm}
\caption{%
Results with different forms of feedback.
}
\label{table:results-comparison-multihf}
% \vspace{-3mm}
\end{table}

\begin{table}[t!]
\centering
% \vspace{-2mm}
\renewcommand{\arraystretch}{1.1} 
\small
% \footnotesize
% \scriptsize
%\begin{tabularx}{1.0\linewidth}{l @{}c HH HH cc cc cccc}
\begin{tabularx}{\linewidth}{l l H HH HH Hc cH Hccc}
\toprule
% & 
% % \multirow{2}{*}{\sc \# Trainable} 
% &
% \multicolumn{2}{c}{ROUGE-1} & 
% \multicolumn{2}{c}{\sc ROUGE-2} &
% % &&&&
% \multicolumn{2}{c}{\sc ROUGE-L} & \\
% \cmidrule(r){3-4} \cmidrule(r){5-6} \cmidrule(r){7-8}
&
% \textsc{Human Feedback Model} 
& 
% \textsc{Parameters} & 
\textbf{\#Params} & 
R@1 & \sc Norm. & %R@1 & 
R@2 & \sc Norm. & %R@2 & 
R@L & 
% \sc Norm. & %R@L &
\textbf{ROUGE-L} &
\textbf{BLEU} & 
\textbf{CIDEr} & 
\textbf{SPICE} &
\textbf{METEOR} &
\\ \midrule

& \textbf{Helpfulness} & 0.25B & 0 & 0 & 0 & 0 & 0 & 0.1520 & 0.0186 & TODO &  0 & 0.1450 \\ 
% & \textbf{Helpfulness} & 0.25B & 0 & 0 & 0 & 0 & 0 & 0.152 & 0.019 & TODO &  0 & 0.145 \\ 
& \textbf{Takeaway} & 0.25B & 0 & 0 & 0 & 0 & 0 &   0.1676 &  0.0230 & 0.000  & 0 & 0.1598 \\ 
& \textbf{Visual} & 0.25B & 0 & 0 & 0 & 0 & 0 &  0.1678 &  0.0230 & 0.0  & 0.0 & 0.1595 \\ 
& \textbf{OCR} & 0.25B & 0 & 0 & 0 & 0 & 0 & 0.1654 & 0.0223 & TODO &  0 & 0.1565 \\ 

% \midrule
% & \textbf{Helpfulness} & 0.25B & 0 & 0 & 0 & 0 & 0 & 0.152 & 0.019 & TODO &  0 & 0.145 \\ 
% & \textbf{Takeaway} & 0.25B & 0 & 0 & 0 & 0 & 0 &   0.168 &  0.023 & 0.000  & 0 & 0.160 \\ 
% & \textbf{Visual} & 0.25B & 0 & 0 & 0 & 0 & 0 &  0.168 &  0.023 & 0.0  & 0.0 & 0.160 \\ 
% & \textbf{OCR} & 0.25B & 0 & 0 & 0 & 0 & 0 & 0.165 & 0.022 & TODO &  0 & 0.157 \\ 

\bottomrule
\end{tabularx}
\vspace{-1mm}
\caption{%
Results with different human feedback metrics.
% (BLIP-RLHF).
}
\label{table:results-comparison-metrics}
\vspace{-3mm}
\end{table}

\begin{table}%[t!]
\centering
\renewcommand{\arraystretch}{1.1} 
\small
% \footnotesize
% \scriptsize
%\begin{tabularx}{1.0\linewidth}{l @{}c HH HH cc cc cccc}
\begin{tabularx}{\linewidth}{l l H HH HH Hc cH Hccc}
\toprule
% & 
% % \multirow{2}{*}{\sc \# Trainable} 
% &
% \multicolumn{2}{c}{ROUGE-1} & 
% \multicolumn{2}{c}{\sc ROUGE-2} &
% % &&&&
% \multicolumn{2}{c}{\sc ROUGE-L} & \\
% \cmidrule(r){3-4} \cmidrule(r){5-6} \cmidrule(r){7-8}
&
% \textsc{Human Feedback Model} 
& 
% \textsc{Parameters} & 
\textbf{\#Params} & 
R@1 & \sc Norm. & %R@1 & 
R@2 & \sc Norm. & %R@2 & 
R@L & 
% \sc Norm. & %R@L &
\textbf{ROUGE-L} &
\textbf{BLEU} & 
\textbf{CIDEr} & 
\textbf{SPICE} &
\textbf{METEOR} &
\\ \midrule

& \textbf{BERT} & 0.25B & 0 & 0 & 0 & 0 & 0 & 0.1565 & 0.01927 & TODO &  0 & 0.1473 \\ 
% & \textbf{Helpfulness} & 0.25B & 0 & 0 & 0 & 0 & 0 & 0.152 & 0.019 & TODO &  0 & 0.145 \\ 
& \textbf{SciBERT} & 0.25B & 0 & 0 & 0 & 0 & 0 &   0.1577 &  0.0201 & 0.000  & 0 & 0.1509 \\ 
& \textbf{BLIP} & 0.25B & 0 & 0 & 0 & 0 & 0 &  0.1573 &  0.01977 & 0.0  & 0.0 & 0.1494 \\

\bottomrule
\end{tabularx}
% \vspace{2mm}
\caption{%
Results with different embedding models for the human-feedback model.
}
\label{table:results-comparison-diff-embedding-models}
\vspace{-4mm}
\end{table}

\begin{table}%[t!]
\centering
\renewcommand{\arraystretch}{1.1} 
\setlength{\tabcolsep}{2pt} 
\small
% \footnotesize
% \scriptsize
%\begin{tabularx}{1.0\linewidth}{l @{}c HH HH cc cc cccc}
\scalebox{0.91}{
\begin{tabularx}{1.1\linewidth}{l l H HH HH Hc cH Hccc}
\toprule
% & 
% % \multirow{2}{*}{\sc \# Trainable} 
% &
% \multicolumn{2}{c}{ROUGE-1} & 
% \multicolumn{2}{c}{\sc ROUGE-2} &
% % &&&&
% \multicolumn{2}{c}{\sc ROUGE-L} & \\
% \cmidrule(r){3-4} \cmidrule(r){5-6} \cmidrule(r){7-8}
&
% \textsc{Human Feedback Model} 
& 
% \textsc{Parameters} & 
\textbf{\#Params} & 
R@1 & \sc Norm. & %R@1 & 
R@2 & \sc Norm. & %R@2 & 
R@L & 
% \sc Norm. & %R@L &
\textbf{ROUGE-L} &
\textbf{BLEU} & 
\textbf{CIDEr} & 
\textbf{SPICE} &
\textbf{METEOR} &
\\ \midrule

& \textbf{BLIP-RLHF} (append) & 0.25B & 0 & 0 & 0 & 0 & 0 & 0.136 & 0.018 & TODO &  0 & 0.132 \\ 
& \textbf{VIT-GPT2-RLHF} (append) & 0.24B & 0 & 0 & 0 & 0 & 0 & 0.138 & 0.016 & TODO &  0 & 0.119 \\ 
& \textbf{BLIP-RLHF} (prepend) & 0.25B & 0 & 0 & 0 & 0 & 0 & 0.152 & \textbf{0.019} & TODO &  0 & \textbf{0.145} \\ 
& \textbf{ViT+GPT2-RLHF} (prepend) & 0.24B & 0 & 0 & 0 & 0 & 0 & 0.138 & \textbf{0.020} & TODO &  0 & 0.126 \\ 

\bottomrule
\end{tabularx}
}
% \vspace{2mm}
\caption{%
Comparing RLHF prepend to append.
}
\label{table:results-comparison-position}
\vspace{-4mm}
\end{table}

%------------------------------------------------------------------------------------
\begin{table}[t!]
\centering
\renewcommand{\arraystretch}{1.1} 
\setlength{\tabcolsep}{7pt} 
\small
\begin{tabular}{l ll l}
\toprule
 && 
 \hspace{15pt}
 \textbf{MSE} \\
\midrule
& Helpfulness & $0.082 \pm 0.12$  & \\
&  Visual & $0.076 \pm 0.20$  \\
& Takeaway & $0.087 \pm 0.17$  \\
& OCR & $0.095 \pm 0.13$  \\
\bottomrule
\end{tabular}
\caption{Evaluation of out-of-sample generalization with respect to different human feedback metrics}
\label{OOS-Gen}
\end{table}
%------------------------------------------------------------------------------------
\begin{table}[ht]
\centering
\vspace{4mm}
\small
\renewcommand{\arraystretch}{1.2} % Adjust row height for better spacing
\begin{tabular}{rcl r c}
\toprule
\textbf{Training Size} & & \textbf{MSE} & \textbf{Gain} & \\
\midrule
25\% (109) && 0.579 & 91.72\% \\
50\% (219) && 0.323 & 6.95\% \\
100\% (438) && 0.311 & 2.98\% \\
125\% (657) && 0.309 & 2.32\% \\
200\% (876) && 0.302 & 0\% \\
\bottomrule
\end{tabular}
\caption{%
Results varying the training size used for learning the human feedback prediction model (for inferring ``Helpfulness''). 
Note gain is computed with respect to the best (lowest) MSE obtained (0.302).
}%

\label{table:varying-training-size}
% \vspace{3mm}
\end{table}

\section{Experiments}\label{sec:exp}

\noindent\textbf{Setup.}
For our human-feedback prediction model, we use MCSE \cite{zhang2022mcse} as embedding function and a 2-layer MLP as regression function. For comparative evaluation, we select the following models as our baselines based on input: (1) OCR-only: Pegasus\cite{zhang2020pegasus}, (2) Figure-only: TrOCR \cite{li2021trocr}, BeiT$+$GPT2, ViT$+$GPT2 \cite{dosovitskiy2020vit}, ViT$+$RoBERTA \cite{dosovitskiy2020vit, liu2019roberta} and (3) Figure-Caption: PromptCap \cite{hu2022promptcap}, Flamingo \cite{alayrac2022flamingo}, GIT \cite{wang2022git}, BLIP \cite{li2022blip} and CLIPCap \cite{mokady2021clipcap}. We use ROUGE-L \cite{lin-2004-rouge}, METEOR \cite{banerjee2005meteor}, BLEU \cite{papineni2002bleu} and CIDEr \cite{vedantam2015cider} metrics for model evaluation. We provide more details regarding individual baselines, metrics, and the dataset in the Appendix.

\subsection{Results}
We show our experimental results in Table ~\ref{table:results-comparison}. We compare our framework with the standard fine-tuning method and benchmark the performance on the Test set of our proposed benchmark. We use BLIP and ViT$+$GPT2 to evaluate our RLHF framework.
From Table ~\ref{table:results-comparison}, we see that models trained using our proposed RLHF formulation perform better than simple fine-tuning.
Specifically, for BLIP, RLHF provides has a 35.7\% increase in BLEU, 16.9\% increase in ROUGE-L, 9\% increase in METEOR and 11.4\% in CIDEr score.
For ViT+GPT2, RLHF provides a 11.1\% increase in BLEU and a 5.1\% increase in CIDEr score.

Aggregating the metrics, we observe that BLIP performs best, which is likely due to its aligned image encoder and text decoder, which are pre-trained jointly. In contrast, ViT+GPT2's modules are not aligned/trained jointly, and the text decoder learns to attend to the vision encoder only during fine-tuning. Thus, improvement with preference alignment is directly related to the choice of the initial pre-trained model. 

% Overall, since the performance increase is generalized among models with different pre-training strategies and overall model-structure, the results show the benefits of using this simple UDRL framework for fine-tuning. Utilizing only a small amount of human annotated data, different scoring mechanisms and prompts can be further developed to take advantage of this limited supervision and further increase performance.  

\subsection{Qualitative Results}
Figure~\ref{fig:HF-predictive-model-top-and-worst-figs} and Figure~\ref{fig:RLHF-vs-base} shows some of the qualitative results of feedback prediction model and the figure-captioning models trained with RLHF. We provide our analysis below:

\textbf{Human Feedback Prediction Model}:
To evaluate the generalizability of our model, we first computed the score predictions on all the of figure-caption pairs. Then we ordered the figure-caption pairs by the predicted scores and selected the top-3 figure-caption pairs with the largest score along with the bottom-3 figure-caption pairs with the lowest score. Results are provided in Figure~\ref{fig:HF-predictive-model-top-and-worst-figs}.
We observe that the figure-caption pairs with the largest scores are highly helpful to the reader (shown in green on the left in Figure~\ref{fig:HF-predictive-model-top-and-worst-figs}), as they mention specific takeaways from the figure (\eg, ``as students make more applications, the number of students who get into their top-choice school decreases, while the number of overall acceptances increases.''), as well as mentioning specific visual aspects that are important to the understanding of the underlying context (\eg, ``... Vertical lines show the true p (blue) and $\beta$ (orange)'').
In contrast, the figure-caption pairs scoring the lowest (bottom-3), which are shown in red on the right in Figure~\ref{fig:HF-predictive-model-top-and-worst-figs}, are vague, without any takeaways, nor reference to visual elements in the figure.

\eat{%}
\rev{%
As an aside, the above results used the helpfulness score from our human experts for training the human feedback prediction model.
However, similar results were observed using the other factors collected from human experts about the figure captions. 
}%
}%

\textbf{Figure-Caption Generative Model}:
From Figure~\ref{fig:RLHF-vs-base} we see that, qualitatively BLIP-RLHF produces better captions compared to fine-tuned BLIP. In most cases, captions produced by BLIP (Fine-tuned) are either explaining the given figure incorrectly (Figure~\ref{fig:RLHF-vs-base}, leftmost sub-figure), not relevant (Figure~\ref{fig:RLHF-vs-base}, middle sub-figure) or are completely uninformative (Figure~\ref{fig:RLHF-vs-base}, rightmost sub-figure). On the other hand, captions produced by BLIP-RLHF method are more faithful to the figure, captures semantic relation between texts to summarize the phenomenon and utilizes visual attributes in explaining the figure. We provide more examples and analysis in the Appendix.

\subsection{Ablation Study} 
We conducted the ablation studies for different components of our framework. We provide our findings below:

\textbf{Effect of granularity of feedback labels}: 
 To evaluate how quantization levels of reward signals (Binary vs. Multi-level) impact model learning, we conducted a comparative study by modifying feedback while training the BLIP-RLHF model.
\\
First, we trained the model for 10 epochs using multi-level human feedback (Row 2), with five feedback levels (very bad, bad, neutral, good, very good) determined at the 20\textsuperscript{th}, 40\textsuperscript{th}, 60\textsuperscript{th}, and 80\textsuperscript{th} percentiles to balance sample distribution. We also experimented with varying label granularity (Row 3), training with binary-label feedback for 5 epochs followed by multi-label feedback for another 5 epochs.
Results in Table~\ref{table:results-comparison-multihf} indicate that both approaches using finer feedback outperform simple binary feedback. Our framework demonstrates the model’s ability to effectively leverage fine-grained feedback. Additionally, the experiment validates the quality of our human prediction model, which provides useful labels at different levels of granularity, enhancing performance for figure-captioning.

\textbf{Comparison of different feedback types}:
To understand the effect of different types of feedback, we compare the results of training the BLIP-RLHF model using Helpfulness, Takeaway, Visual-descriptiveness (Visual), and Image-text (OCR) feedback scores. The results are provided in Table~\ref{table:results-comparison-metrics}.
We observe that training BLIP-RLHF with Takeaway, Visual, and OCR feedback outperforms training with Helpfulness feedback. This is expected, as the Helpfulness rating is subjective, whereas Visual and Takeaway are objective evaluation metrics. This finding highlights the importance of feedback type and suggests that further improvements can be achieved by modeling different aspects of the annotated human dataset.

\textbf{Feedback prediction model architecture}:

We compare different embedding models (BERT, SciBERT and BLIP) in constructing the human feedback prediction model. The results are provided in Table~\ref{table:results-comparison-diff-embedding-models}. 
We observe that different representations outperform our default MCSE implementation, indicating that our human feedback prediction model, and downstream figure-captioning performance, are sensitive to the quality of representations used. This highlights that, further performance gains can be made by using different representations, for example, by encoding different modalities (text only vs joint encoding of text and vision).

\textbf{Generalizability of the human feedback prediction model}: To evaluate the out-of-sample generalization of our human-feedback prediction model, we conduct a 5-fold cross-validation experiment on the original 438 annotated.
We repeated the above experiment 5 times. 

We report our results in Table \ref{OOS-Gen}, including mean squared error (MSE) and standard deviation. 

As can be seen from Table~\ref{OOS-Gen}, our model is able to achieve good results on the validation set. 

This highlights that our human-feedback prediction model demonstrates out-of-sample generalization and proves the statistical significance of our model.

\textbf{Varying training size}:
To evaluate the effectiveness of our approach when varying the number of samples used during training, we train the human feedback prediction model using 25\%, 50\%, 100\%, 125\%, and 200\% of the human-annotated data.
We used a held-out set of 300 samples for model evaluation of each of these models.
We then trained separate models for each training set for the task of predicting the 'Helpfulness' measure. 
The results showing mean-squared error (MSE; lower is better) are provided in Table~\ref{table:varying-training-size}.
Notably, we see the test performance of the model saturates as the number of training samples is increased. Even with 50\% of the original human-annotated data, the model achieves good test results.

\textbf{Effect of human feedback position}: 
To understand the sensitivity of the model to the position of human feedback, we compare the performance of appending and pre-pending the human feedback labels in Table ~\ref{table:results-comparison-position}. Since our models generate text, during test time, without any human feedback label prompt, they can only rely on feedback during training. Additionally, due to the auto-regressive generation of our models, they only observe the label before generation, and for append, only observe the label after generation. Intuitively, pre-pending should work best since the generation is conditioned on the label. The results support this and show that ViT+GPT2 and BLIP perform better when trained with pre-pended human feedback.

%======================================================================================

\section{Conclusion} \label{sec:conc}

In this work, we developed a new benchmark and methodology to improve caption generation for scientific figures. 

We showed that incorporating domain expert feedback in learning a model for figure-to-caption generation improves both model performance and caption quality.

Our proposed framework is scalable (requires limited manual human effort in labeling) and flexible (allows for incorporating multiple reward signals at different granularity). 
% The proposed benchmark of figure-caption pairs with caption quality scores to further the research efforts in reader-aligned figure-captioning tasks. 
% We also propose a new benchmark of figure-caption pairs with caption quality scores to further the research efforts in reader-aligned figure-captioning tasks. 
We hope that this new benchmark dataset will allow researchers to benchmark their own methods for incorporating human feedback in figure-to-caption generation tasks and various other image-to-text generation tasks.

Future work will explore techniques to incorporate multiple complementary feedback as well as different ways to quantize the reward score to leverage it as valid feedback when training the model.

\section*{Limitations}
\label{sec:limit}
Our work, while improving scientific figure caption generation with respect to general reader preferences, still has certain limitations, which require further considerations:

\textbf{Fine-tuning of Caption generation model:} Our UDRL-based fine-tuning scheme for training caption generation models currently requires us to update all the parameters of the model. This can lead to the usage of more compute resources when compared to methods like Parameter Efficient Fine-Tuning algorithms, like the Low-Rank Adaptation method.

\textbf{Feedback annotation for specific reader groups:} Our work currently focuses on improving model alignment with respect to general readers. However, an important use case of automatic figure-captioning is building accessible assistive tools for specific users groups, for example, people with visual impairments. This requires additional consideration when generating initial reader-preference feedback annotations. 
% }

%----------------------------------------------------------
\bibliography{main}

\begin{thebibliography}{51}
\expandafter\ifx\csname natexlab\endcsname\relax\def\natexlab#1{#1}\fi

\bibitem[{Alayrac et~al.(2022)Alayrac, Donahue, Luc, Miech, Barr, Hasson, Lenc, Mensch, Millican, Reynolds et~al.}]{alayrac2022flamingo}
Jean-Baptiste Alayrac, Jeff Donahue, Pauline Luc, Antoine Miech, Iain Barr, Yana Hasson, Karel Lenc, Arthur Mensch, Katie Millican, Malcolm Reynolds, et~al. 2022.
\newblock Flamingo: a visual language model for few-shot learning.
\newblock \emph{arXiv preprint arXiv:2204.14198}.

\bibitem[{Bahdanau et~al.(2016)Bahdanau, Brakel, Xu, Goyal, Lowe, Pineau, Courville, and Bengio}]{bahdanau2016actor}
Dzmitry Bahdanau, Philemon Brakel, Kelvin Xu, Anirudh Goyal, Ryan Lowe, Joelle Pineau, Aaron Courville, and Yoshua Bengio. 2016.
\newblock An actor-critic algorithm for sequence prediction.
\newblock \emph{arXiv preprint arXiv:1607.07086}.

\bibitem[{Banerjee and Lavie(2005)}]{banerjee2005meteor}
Satanjeev Banerjee and Alon Lavie. 2005.
\newblock Meteor: An automatic metric for mt evaluation with improved correlation with human judgments.
\newblock In \emph{Proceedings of the acl workshop on intrinsic and extrinsic evaluation measures for machine translation and/or summarization}, pages 65--72.

\bibitem[{Bao et~al.(2022)Bao, Dong, Piao, and Wei}]{bao2022beit}
Hangbo Bao, Li~Dong, Songhao Piao, and Furu Wei. 2022.
\newblock \href {https://openreview.net/forum?id=p-BhZSz59o4} {{BE}it: {BERT} pre-training of image transformers}.
\newblock In \emph{International Conference on Learning Representations}.

\bibitem[{Chaudhry et~al.(2020)Chaudhry, Shekhar, Gupta, Maneriker, Bansal, and Joshi}]{chaudhry2020leaf}
Ritwick Chaudhry, Sumit Shekhar, Utkarsh Gupta, Pranav Maneriker, Prann Bansal, and Ajay Joshi. 2020.
\newblock Leaf-qa: Locate, encode \& attend for figure question answering.
\newblock In \emph{Proceedings of the IEEE/CVF Winter Conference on Applications of Computer Vision}, pages 3512--3521.

\bibitem[{Chen et~al.(2019)Chen, Zhang, Kim, Cohen, Yu, Rossi, and Bunescu}]{chen2019neural}
Charles Chen, Ruiyi Zhang, Sungchul Kim, Scott Cohen, Tong Yu, Ryan Rossi, and Razvan Bunescu. 2019.
\newblock Neural caption generation over figures.
\newblock In \emph{Adjunct Proceedings of the 2019 ACM International Joint Conference on Pervasive and Ubiquitous Computing and Proceedings of the 2019 ACM International Symposium on Wearable Computers}, pages 482--485.

\bibitem[{Chen et~al.(2020{\natexlab{a}})Chen, Zhang, Koh, Kim, Cohen, and Rossi}]{chen2019figure}
Charles Chen, Ruiyi Zhang, Eunyee Koh, Sungchul Kim, Scott Cohen, and Ryan Rossi. 2020{\natexlab{a}}.
\newblock \href {https://doi.org/10.1109/WACV45572.2020.9093592} {Figure captioning with relation maps for reasoning}.
\newblock In \emph{2020 IEEE Winter Conference on Applications of Computer Vision (WACV)}, pages 1526--1534.

\bibitem[{Chen et~al.(2020{\natexlab{b}})Chen, Zhang, Koh, Kim, Cohen, and Rossi}]{chen2020figure}
Charles Chen, Ruiyi Zhang, Eunyee Koh, Sungchul Kim, Scott Cohen, and Ryan Rossi. 2020{\natexlab{b}}.
\newblock Figure captioning with relation maps for reasoning.
\newblock In \emph{Proceedings of the IEEE/CVF Winter Conference on Applications of Computer Vision}, pages 1537--1545.

\bibitem[{Devlin et~al.(2018)Devlin, Chang, Lee, and Toutanova}]{devlin2018bert}
Jacob Devlin, Ming-Wei Chang, Kenton Lee, and Kristina Toutanova. 2018.
\newblock Bert: Pre-training of deep bidirectional transformers for language understanding.
\newblock \emph{arXiv preprint arXiv:1810.04805}.

\bibitem[{Dosovitskiy et~al.(2021)Dosovitskiy, Beyer, Kolesnikov, Weissenborn, Zhai, Unterthiner, Dehghani, Minderer, Heigold, Gelly, Uszkoreit, and Houlsby}]{dosovitskiy2020vit}
Alexey Dosovitskiy, Lucas Beyer, Alexander Kolesnikov, Dirk Weissenborn, Xiaohua Zhai, Thomas Unterthiner, Mostafa Dehghani, Matthias Minderer, Georg Heigold, Sylvain Gelly, Jakob Uszkoreit, and Neil Houlsby. 2021.
\newblock An image is worth 16x16 words: Transformers for image recognition at scale.
\newblock \emph{ICLR}.

\bibitem[{Emmons et~al.(2021)Emmons, Eysenbach, Kostrikov, and Levine}]{emmons2021rvs}
Scott Emmons, Benjamin Eysenbach, Ilya Kostrikov, and Sergey Levine. 2021.
\newblock Rvs: What is essential for offline rl via supervised learning?
\newblock \emph{arXiv preprint arXiv:2112.10751}.

\bibitem[{Gebru et~al.(2021)Gebru, Morgenstern, Vecchione, Vaughan, Wallach, Iii, and Crawford}]{gebru2021datasheets}
Timnit Gebru, Jamie Morgenstern, Briana Vecchione, Jennifer~Wortman Vaughan, Hanna Wallach, Hal~Daum{\'e} Iii, and Kate Crawford. 2021.
\newblock Datasheets for datasets.
\newblock \emph{Communications of the ACM}, 64(12):86--92.

\bibitem[{Hsu et~al.(2021)Hsu, Giles, and Huang}]{hsu-etal-2021-scicap-generating}
Ting-Yao Hsu, C~Lee Giles, and Ting-Hao Huang. 2021.
\newblock \href {https://doi.org/10.18653/v1/2021.findings-emnlp.277} {{S}ci{C}ap: Generating captions for scientific figures}.
\newblock In \emph{Findings of the Association for Computational Linguistics: EMNLP 2021}, pages 3258--3264, Punta Cana, Dominican Republic. Association for Computational Linguistics.

\bibitem[{Hu et~al.(2022)Hu, Hua, Yang, Shi, Smith, and Luo}]{hu2022promptcap}
Yushi Hu, Hang Hua, Zhengyuan Yang, Weijia Shi, Noah~A Smith, and Jiebo Luo. 2022.
\newblock Promptcap: Prompt-guided task-aware image captioning.
\newblock \emph{arXiv:2211.09699}.

\bibitem[{Huang et~al.(2023)}]{summaries-as-captions-preprint}
Chieh-Yang Huang et~al. 2023.
\newblock Summaries as captions: Generating figure captions for scientific documents with automated text summarization.
\newblock \emph{Open Review}.
\newblock \url{https://openreview.net/pdf?id=80R7RVLcsf}.

\bibitem[{Ibarz et~al.(2018)Ibarz, Leike, Pohlen, Irving, Legg, and Amodei}]{ibarz2018reward}
Borja Ibarz, Jan Leike, Tobias Pohlen, Geoffrey Irving, Shane Legg, and Dario Amodei. 2018.
\newblock Reward learning from human preferences and demonstrations in atari.
\newblock \emph{Advances in neural information processing systems}, 31.

\bibitem[{Kafle et~al.(2018)Kafle, Price, Cohen, and Kanan}]{kafle2018dvqa}
Kushal Kafle, Brian Price, Scott Cohen, and Christopher Kanan. 2018.
\newblock Dvqa: Understanding data visualizations via question answering.
\newblock In \emph{Proceedings of the IEEE conference on computer vision and pattern recognition}, pages 5648--5656.

\bibitem[{Kafle et~al.(2020)Kafle, Shrestha, Cohen, Price, and Kanan}]{kafle2020answering}
Kushal Kafle, Robik Shrestha, Scott Cohen, Brian Price, and Christopher Kanan. 2020.
\newblock Answering questions about data visualizations using efficient bimodal fusion.
\newblock In \emph{Proceedings of the IEEE/CVF Winter conference on applications of computer vision}, pages 1498--1507.

\bibitem[{Kahou et~al.(2017)Kahou, Michalski, Atkinson, K{\'a}d{\'a}r, Trischler, and Bengio}]{kahou2017figureqa}
Samira~Ebrahimi Kahou, Vincent Michalski, Adam Atkinson, {\'A}kos K{\'a}d{\'a}r, Adam Trischler, and Yoshua Bengio. 2017.
\newblock Figureqa: An annotated figure dataset for visual reasoning.
\newblock \emph{Sixth International Conference on Learning Representations Workshop}.

\bibitem[{Korbak et~al.(2023)Korbak, Shi, Chen, Bhalerao, Buckley, Phang, Bowman, and Perez}]{korbak2023pretraining}
Tomasz Korbak, Kejian Shi, Angelica Chen, Rasika Bhalerao, Christopher~L Buckley, Jason Phang, Samuel~R Bowman, and Ethan Perez. 2023.
\newblock Pretraining language models with human preferences.
\newblock \emph{arXiv preprint arXiv:2302.08582}.

\bibitem[{Kreutzer et~al.(2018)Kreutzer, Khadivi, Matusov, and Riezler}]{kreutzer2018can}
Julia Kreutzer, Shahram Khadivi, Evgeny Matusov, and Stefan Riezler. 2018.
\newblock Can neural machine translation be improved with user feedback?
\newblock \emph{arXiv preprint arXiv:1804.05958}.

\bibitem[{Lee et~al.(2023)Lee, Liu, Ryu, Watkins, Du, Boutilier, Abbeel, Ghavamzadeh, and Gu}]{lee2023aligning}
Kimin Lee, Hao Liu, Moonkyung Ryu, Olivia Watkins, Yuqing Du, Craig Boutilier, Pieter Abbeel, Mohammad Ghavamzadeh, and Shixiang~Shane Gu. 2023.
\newblock Aligning text-to-image models using human feedback.
\newblock \emph{arXiv preprint arXiv:2302.12192}.

\bibitem[{Lee et~al.(2021)Lee, Smith, and Abbeel}]{lee2021pebble}
Kimin Lee, Laura Smith, and Pieter Abbeel. 2021.
\newblock Pebble: Feedback-efficient interactive reinforcement learning via relabeling experience and unsupervised pre-training.
\newblock \emph{arXiv preprint arXiv:2106.05091}.

\bibitem[{Li et~al.(2022{\natexlab{a}})Li, Li, Xiong, and Hoi}]{li2022blip}
Junnan Li, Dongxu Li, Caiming Xiong, and Steven Hoi. 2022{\natexlab{a}}.
\newblock Blip: Bootstrapping language-image pre-training for unified vision-language understanding and generation.
\newblock In \emph{International Conference on Machine Learning}, pages 12888--12900. PMLR.

\bibitem[{Li et~al.(2021)Li, Lv, Cui, Lu, Florencio, Zhang, Li, and Wei}]{li2021trocr}
Minghao Li, Tengchao Lv, Lei Cui, Yijuan Lu, Dinei Florencio, Cha Zhang, Zhoujun Li, and Furu Wei. 2021.
\newblock Trocr: Transformer-based optical character recognition with pre-trained models.
\newblock \emph{arXiv preprint arXiv:2109.10282}.

\bibitem[{Li et~al.(2022{\natexlab{b}})Li, Wu, and Chen}]{li2022multi}
Ying Li, Qingfeng Wu, and Bin Chen. 2022{\natexlab{b}}.
\newblock Multi-attention relation network for figure question answering.
\newblock In \emph{Knowledge Science, Engineering and Management: 15th International Conference, KSEM 2022, Singapore, August 6--8, 2022, Proceedings, Part II}, pages 667--680. Springer.

\bibitem[{Lin(2004)}]{lin-2004-rouge}
Chin-Yew Lin. 2004.
\newblock \href {https://aclanthology.org/W04-1013} {{ROUGE}: A package for automatic evaluation of summaries}.
\newblock In \emph{Text Summarization Branches Out}, pages 74--81, Barcelona, Spain. Association for Computational Linguistics.

\bibitem[{Liu et~al.(2019)Liu, Ott, Goyal, Du, Joshi, Chen, Levy, Lewis, Zettlemoyer, and Stoyanov}]{liu2019roberta}
Yinhan Liu, Myle Ott, Naman Goyal, Jingfei Du, Mandar Joshi, Danqi Chen, Omer Levy, Mike Lewis, Luke Zettlemoyer, and Veselin Stoyanov. 2019.
\newblock Roberta: A robustly optimized bert pretraining approach.
\newblock \emph{arXiv preprint arXiv:1907.11692}.

\bibitem[{Lu et~al.(2022)Lu, Welleck, Hessel, Jiang, Qin, West, Ammanabrolu, and Choi}]{lu2022quark}
Ximing Lu, Sean Welleck, Jack Hessel, Liwei Jiang, Lianhui Qin, Peter West, Prithviraj Ammanabrolu, and Yejin Choi. 2022.
\newblock Quark: Controllable text generation with reinforced unlearning.
\newblock \emph{Advances in neural information processing systems}, 35:27591--27609.

\bibitem[{MacGlashan et~al.(2017)MacGlashan, Ho, Loftin, Peng, Wang, Roberts, Taylor, and Littman}]{macglashan2017interactive}
James MacGlashan, Mark~K Ho, Robert Loftin, Bei Peng, Guan Wang, David~L Roberts, Matthew~E Taylor, and Michael~L Littman. 2017.
\newblock Interactive learning from policy-dependent human feedback.
\newblock In \emph{International Conference on Machine Learning}, pages 2285--2294. PMLR.

\bibitem[{Masry et~al.(2022)Masry, Long, Tan, Joty, and Hoque}]{masry2022chartqa}
Ahmed Masry, Do~Xuan Long, Jia~Qing Tan, Shafiq Joty, and Enamul Hoque. 2022.
\newblock Chartqa: A benchmark for question answering about charts with visual and logical reasoning.
\newblock \emph{arXiv preprint arXiv:2203.10244}.

\bibitem[{Methani et~al.(2020)Methani, Ganguly, Khapra, and Kumar}]{methani2020plotqa}
Nitesh Methani, Pritha Ganguly, Mitesh~M Khapra, and Pratyush Kumar. 2020.
\newblock Plotqa: Reasoning over scientific plots.
\newblock In \emph{Proceedings of the IEEE/CVF Winter Conference on Applications of Computer Vision}, pages 1527--1536.

\bibitem[{Mokady et~al.(2021)Mokady, Hertz, and Bermano}]{mokady2021clipcap}
Ron Mokady, Amir Hertz, and Amit~H Bermano. 2021.
\newblock Clipcap: Clip prefix for image captioning.
\newblock \emph{arXiv preprint arXiv:2111.09734}.

\bibitem[{Papineni et~al.(2002)Papineni, Roukos, Ward, and Zhu}]{papineni2002bleu}
Kishore Papineni, Salim Roukos, Todd Ward, and Wei-Jing Zhu. 2002.
\newblock Bleu: a method for automatic evaluation of machine translation.
\newblock In \emph{Proceedings of the 40th annual meeting of the Association for Computational Linguistics}, pages 311--318.

\bibitem[{Qian et~al.(2020)Qian, Koh, Du, Kim, and Chan}]{qian2020formative}
Xin Qian, Eunyee Koh, Fan Du, Sungchul Kim, and Joel Chan. 2020.
\newblock A formative study on designing accurate and natural figure captioning systems.
\newblock In \emph{Extended Abstracts of the 2020 CHI Conference on Human Factors in Computing Systems}, pages 1--8.

\bibitem[{Qian et~al.(2021)Qian, Koh, Du, Kim, Chan, Rossi, Malik, and Lee}]{capunits2021}
Xin Qian, Eunyee Koh, Fan Du, Sungchul Kim, Joel Chan, Ryan~A. Rossi, Sana Malik, and Tak~Yeon Lee. 2021.
\newblock \href {https://doi.org/10.1145/3442381.3449923} {Generating accurate caption units for figure captioning}.
\newblock In \emph{Proceedings of the Web Conference 2021}, WWW '21, page 2792–2804, New York, NY, USA. Association for Computing Machinery.

\bibitem[{Radford et~al.(2019)Radford, Wu, Child, Luan, Amodei, Sutskever et~al.}]{radford2019language}
Alec Radford, Jeffrey Wu, Rewon Child, David Luan, Dario Amodei, Ilya Sutskever, et~al. 2019.
\newblock Language models are unsupervised multitask learners.
\newblock \emph{OpenAI blog}, 1(8):9.

\bibitem[{Roberts et~al.(2024)Roberts, Han, Houlsby, and Albanie}]{roberts2024scifibench}
Jonathan Roberts, Kai Han, Neil Houlsby, and Samuel Albanie. 2024.
\newblock Scifibench: Benchmarking large multimodal models for scientific figure interpretation.
\newblock \emph{arXiv preprint arXiv:2405.08807}.

\bibitem[{Schulman et~al.(2017)Schulman, Wolski, Dhariwal, Radford, and Klimov}]{schulman2017proximal}
John Schulman, Filip Wolski, Prafulla Dhariwal, Alec Radford, and Oleg Klimov. 2017.
\newblock Proximal policy optimization algorithms.
\newblock \emph{arXiv preprint arXiv:1707.06347}.

\bibitem[{Siegel et~al.(2016)Siegel, Horvitz, Levin, Divvala, and Farhadi}]{siegel2016figureseer}
Noah Siegel, Zachary Horvitz, Roie Levin, Santosh Divvala, and Ali Farhadi. 2016.
\newblock Figureseer: Parsing result-figures in research papers.
\newblock In \emph{European Conference on Computer Vision}, pages 664--680. Springer.

\bibitem[{Singh and Shekhar(2020)}]{singh2020stl}
Hrituraj Singh and Sumit Shekhar. 2020.
\newblock Stl-cqa: Structure-based transformers with localization and encoding for chart question answering.
\newblock In \emph{Proceedings of the 2020 Conference on Empirical Methods in Natural Language Processing (EMNLP)}, pages 3275--3284.

\bibitem[{Srivastava et~al.(2019)Srivastava, Shyam, Mutz, Ja{\'s}kowski, and Schmidhuber}]{srivastava2019training}
Rupesh~Kumar Srivastava, Pranav Shyam, Filipe Mutz, Wojciech Ja{\'s}kowski, and J{\"u}rgen Schmidhuber. 2019.
\newblock Training agents using upside-down reinforcement learning.
\newblock \emph{arXiv preprint arXiv:1912.02877}.

\bibitem[{Stefanini et~al.(2022)Stefanini, Cornia, Baraldi, Cascianelli, Fiameni, and Cucchiara}]{stefanini2022show}
Matteo Stefanini, Marcella Cornia, Lorenzo Baraldi, Silvia Cascianelli, Giuseppe Fiameni, and Rita Cucchiara. 2022.
\newblock From show to tell: a survey on deep learning-based image captioning.
\newblock \emph{IEEE transactions on pattern analysis and machine intelligence}, 45(1):539--559.

\bibitem[{Stiennon et~al.(2020)Stiennon, Ouyang, Wu, Ziegler, Lowe, Voss, Radford, Amodei, and Christiano}]{stiennon2020learning}
Nisan Stiennon, Long Ouyang, Jeffrey Wu, Daniel Ziegler, Ryan Lowe, Chelsea Voss, Alec Radford, Dario Amodei, and Paul~F Christiano. 2020.
\newblock Learning to summarize with human feedback.
\newblock \emph{Advances in Neural Information Processing Systems}, 33:3008--3021.

\bibitem[{Vedantam et~al.(2015)Vedantam, Lawrence~Zitnick, and Parikh}]{vedantam2015cider}
Ramakrishna Vedantam, C~Lawrence~Zitnick, and Devi Parikh. 2015.
\newblock Cider: Consensus-based image description evaluation.
\newblock In \emph{Proceedings of the IEEE conference on computer vision and pattern recognition}, pages 4566--4575.

\bibitem[{Wang et~al.(2022{\natexlab{a}})Wang, Yang, Hu, Li, Lin, Gan, Liu, Liu, and Wang}]{wang2022git}
Jianfeng Wang, Zhengyuan Yang, Xiaowei Hu, Linjie Li, Kevin Lin, Zhe Gan, Zicheng Liu, Ce~Liu, and Lijuan Wang. 2022{\natexlab{a}}.
\newblock Git: A generative image-to-text transformer for vision and language.
\newblock \emph{arXiv preprint arXiv:2205.14100}.

\bibitem[{Wang et~al.(2022{\natexlab{b}})Wang, Yang, Men, Lin, Bai, Li, Ma, Zhou, Zhou, and Yang}]{wang2022ofa}
Peng Wang, An~Yang, Rui Men, Junyang Lin, Shuai Bai, Zhikang Li, Jianxin Ma, Chang Zhou, Jingren Zhou, and Hongxia Yang. 2022{\natexlab{b}}.
\newblock Ofa: Unifying architectures, tasks, and modalities through a simple sequence-to-sequence learning framework.
\newblock In \emph{International Conference on Machine Learning}, pages 23318--23340. PMLR.

\bibitem[{Zhang et~al.(2020)Zhang, Zhao, Saleh, and Liu}]{zhang2020pegasus}
Jingqing Zhang, Yao Zhao, Mohammad Saleh, and Peter Liu. 2020.
\newblock Pegasus: Pre-training with extracted gap-sentences for abstractive summarization.
\newblock In \emph{International Conference on Machine Learning}, pages 11328--11339. PMLR.

\bibitem[{Zhang et~al.(2022)Zhang, Mosbach, Adelani, Hedderich, and Klakow}]{zhang2022mcse}
Miaoran Zhang, Marius Mosbach, David~Ifeoluwa Adelani, Michael~A Hedderich, and Dietrich Klakow. 2022.
\newblock Mcse: Multimodal contrastive learning of sentence embeddings.
\newblock \emph{arXiv preprint arXiv:2204.10931}.

\bibitem[{Zhang et~al.(2023)Zhang, Yang, Feng, Qin, Chen, Yu, Chen, Wang, Savarese, Ermon et~al.}]{zhang2023hive}
Shu Zhang, Xinyi Yang, Yihao Feng, Can Qin, Chia-Chih Chen, Ning Yu, Zeyuan Chen, Huan Wang, Silvio Savarese, Stefano Ermon, et~al. 2023.
\newblock Hive: Harnessing human feedback for instructional visual editing.
\newblock \emph{arXiv preprint arXiv:2303.09618}.

\bibitem[{Zou et~al.(2020)Zou, Wu, Xue, and Wu}]{zou2020affinity}
Jialong Zou, Guoli Wu, Taofeng Xue, and Qingfeng Wu. 2020.
\newblock An affinity-driven relation network for figure question answering.
\newblock In \emph{2020 IEEE International Conference on Multimedia and Expo (ICME)}, pages 1--6. IEEE.

\end{thebibliography}

\appendix

\section*{Appendix}
\label{sec:appendix}

\section{Overview}

In the following subsections, 

\begin{itemize}
    \item 
    % This document (supplementary.pdf): 
    We provide details of our quality metrics used for evaluating a figure-caption pair, our experimental setup, baseline model details and a discussion on the qualitative comparitive results.
    \item 
    % datasheet.pdf : 
    Following the guidelines mentioned in \cite{gebru2021datasheets}, we provide information regarding data composition, data collection procedure, use cases for our dataset. The document also includes Author statement, Licensing and Maintenance Plan.
\end{itemize}

Our dataset along with its documentation and code has been made publicly available at:
% . The links for same are as follows:
\\
% \small
\textbf{Benchmark: }\href{https://figshare.com/s/c034fd77bea9475319cb}{Benchmark Link}\\
% \small
\textbf{Code: }\href{https://github.com/FigCapsHF/FigCapsHF}{Codebase Link}\\
% \small
\textbf{Documentation: }\href{https://figcapshf.github.io/}{Documentation Link}\\

\subsection{Ethics Statement} \label{sec:eth}
Our work on improving figure caption generation is important in building accessible assistive tools for the scientific community. However, like many works in the area of generative AI, our work/general ideas also carry the risk of misuse i.e. our proposed method can be advertised by a third party as a deployable product, when in fact, we believe that our proposed method is a research endeavor and still has room for improvement. Another potential negative impact of our work could be the complacent consideration of generating human feedback without due consideration to human subjects involved. This is our key motivation to make our dataset with feedback labels public, to allow interested researchers to develop and benchmark their own methods that require feedback.

Finally, we comment on the dataset privacy considerations for the proposed benchmark. Our proposed dataset and other datasets considered in this work are licensed for academic/non-commercial research (Creative Commons Attribution-Non Commercial-Share Alike 4.0 International License). Our proposed dataset does not contain any personal information.

 \subsection{Description of metrics used for Feedback assessment}

We followed \cite{summaries-as-captions-preprint} to evaluate a given figure-caption pair from the perspective of a reader. Specifically, we used the following measures:

\begin{itemize}
    \item \textbf{Helpfulness:} This is a subjective measure to evaluate whether a given caption is able to inform the reader about the information conveyed in the corresponding figure.
    \item \textbf{Takeaway:} This measure is used to assess a given caption based on whether it is able to convey a conclusive information about the given figure image. 
    \item \textbf{Visual-descriptiveness (visual):} We define visual descriptiveness of a given caption as a measure of how much the given caption is grounded with respect to the figure. For example, a caption that describes the visual elements of the figure like color and shape should be more informative to the readers.
    \item \textbf{Image-text (OCR):} We formulate OCR as a metric to evaluate if the given caption included textual elements of the figure like title, legends and labels when describing the figure.
\end{itemize}

\subsection{Experimental Setup}

\subsubsection{Datasets}
\label{sec:data}
For all our models, we use the same splits in our benchmark dataset; this portion contains 106,834 training pairs, 13,354 validation pairs, and 13,355 test pairs. The primary difference between our baseline and RLHF models is the human-feedback augmented figure-captions that are used for training the latter (figure-images remain the same) and testing figure-caption pairs remain the same for both.

\textbf{Annotation details of Human-Feedback set}: 
We selected the annotators based on their expertise in the areas of computer vision/natural language processing and machine learning. Our annotator pool consisted of 10 Ph.D. graduates and active graduate students (no authors) with published work in the CV, NLP, and ML conferences. We randomly selected 438 figure-caption pairs from the dataset to be annotated. Each annotator was provided 2 weeks time to annotate the data subset. For each sample, annotators were asked to provide ratings on a five-point Likert scale for the following attributes [OCR, Visual, Takeaway, Helpfulness]. For each sample, the following descriptions were provided:

\begin{itemize}
    \item OCR: The caption includes named entities or important words/numbers in the figure(e.g., title, legends, labels, etc.).
    \item Visual-Descriptiveness: The caption includes some visual characteristics of the figure (e.g., color, shape, trend, etc.). 
    \item Takeaway: The given caption explicitly states the high-level takeaway message or the conclusion that the figure attempted to convey. 
    \item Helpfulness: The caption was helpful in understanding the message that the figure is attempting to convey.
\end{itemize}

\textbf{Human-Feedback Augmented Caption}
For our RLHF-trained models, we generate human-feedback augmented figure-captions to align the model to human preferences. 
In this process, for each caption, we first use MCSE \cite{zhang2022mcse} to generate text-embeddings for the captions in the human annotated dataset (~400 pairs). An auxiliary scoring-model (MLP Regressor) is then trained to predict the reader-preference scores using these embeddings, and later used to predict human feedback scores for the entire dataset; we pick the median of these scores as a pivot and label all captions with higher scores as "good", and lower scores as "bad". After pre-pending our captions with these annotations, we effectively train our models in a UDRL framework. Code to implement and generate new human-feedback augmented captions are provided in the GitHub repository. 

\subsubsection{Evaluation Metrics}
We evaluate the generated captions using a variety of common metrics. 
\textbf{ROUGE-L} \cite{lin-2004-rouge} is a recall-oriented metric which uses the Longest Common Subsequence between the reference and the model generated caption, we report the F1 score. 
\textbf{BLEU}\cite{papineni2002bleu} is a precision-oriented metric which uses n-gram overlap, and an additional penalty for sentence brevity. Here, we are using \textbf{BLEU@4} (i.e $n = 4$ for n-gram overlap)
\textbf{METEOR}\cite{banerjee2005meteor} measures generalized unigram-overlap and computes a combination of the precision and recall. 
For a summary of the evaluation metrics leveraged by traditional image captioning works, see~\cite{stefanini2022show}.

\subsubsection{Baselines}
% %input,output,optimizer,hyperparameter, dataset, finetuning/pretraining strategy, number of epochs, learning rate,text generation strategy, learning rate scheduler, batch size, versions of the models specifically

For comparative evaluation of our proposed framework, we selected methods based on the information used to generate a caption. Specifically, we categorize the baselines models into following categories: 
\begin{itemize}
    \item \textbf{Figure-only:} We refer a method as 'Figure-only' if the given method computes an output text based on uni-modal embedding of the input image. Model architecture under this category generally comprises of some combination of a vision encoder and a text decoder module.  
    \item \textbf{OCR-only:} Similar to above, if a method generates an output text using only text as input to the text decoder model, we classify the same as 'Text-only' methods. Specific to our case, we can extract some textual descriptions of a given figure by applying an off-the-shelf OCR method. Hence from here on , w would explicitly refer to methods falling under the above mentioned criteria as 'OCR-only' models. Methods under this category utilizes a text encoder and text decoder modules as part of their model architecture. 
    \item \textbf{Figure-Caption:} Finally for methods which compute multi-modal embedding from text and image uni-modal embeddings to be utilized for generating output text using a text decoder, we categorize  them as 'Figure-Caption' methods. All the methods under this category generally include a vision encoder, text encoder and text decoder modules as part of their model architecture.
\end{itemize} 

We evaluate a variety of strong image-captioning models and a text-summarization model as our baselines. We provide details of individual models below:

\textbf{Unimodal Vision-Encoder Language-Decoder Models}. These models consist of a pre-trained Vision-Encoder (e.g. BEiT \cite{bao2022beit}, ViT \cite{dosovitskiy2020vit}) and a pre-trained Text-Decoder/Language model (e.g. GPT-2 \cite{radford2019language}, RoBERTA \cite{liu2019roberta}). The two submodules are not pre-trained jointly, and only aligned during fine-tuning via randomly initialized cross-attention layers in the decoder. These models simply take in the figure-image and generate the corresponding caption.
\\
\textbf{Pegasus} \cite{zhang2020pegasus} is a Transformer-based pre-trained model for text-summarization. We use PEGASUS to generate figure-captions by summarizing the OCR extracted from the image itself. 
\\
\textbf{TrOCR} \cite{li2021trocr} is a Transformer-based OCR model designed to extract text from a given image. It uses BEiT/DEiT as a vision encoder and RoBERTA as a text decoder, similar to the aforementioned image-to-text models, with the addition of an OCR-focused pre-training. We fine-tuned the model to generate a caption from a given figure-image. 
\\
\textbf{GIT} \cite{wang2022git} is a Generative Image-to-Text model. It uses a pre-trained Vision-Transformer encoder and a randomly initialized Language Transformer decoder (e.g. BERT\cite{devlin2018bert}), similar to the aforementioned image-to-text models, and further jointly pre-trains them using the Language Modeling task. We evaluated the performance of both fine-tuned and pre-trained versions of GIT.  
\\
\textbf{BLIP} \cite{li2022blip} is a Multi-Modal Vision-Language decoder model. It has a similar architecture to the Vision-Encoder Decoder image-to-text models, but utilizes interchangeable attention layers in the text-decoder to behave as either an unimodal encoder, an image-grounded text encoder or an image-grounded text decoder. The model is pre-trained using the LM, ITM and ITC losses jointly. 
\\
\textbf{PromptCap}\cite{hu2022promptcap} is a prompt-based image-captioning model. In addition to taking an image, the model can also incorporates a user-defined prompt to guide the generated caption. PromptCap  utilizes a pre-trained Transformer-based encoder-decoder model, namely OFA \cite{wang2022ofa} which is further pre-trained. PromptCap is evaluated zero-shot using its pre-trained version due to lack of available documentation. 
\\
\textbf{Flamingo-mini} \cite{alayrac2022flamingo} is a Transformer-based encoder-decoder model which has a similar structure to the aforementioned image-to-text models. However, the pre-trained vision encoder and text decoder are frozen and an additional module is used to learn transformed visual representations for the frozen language model to attend to. 
\\
\textbf{CLIPCap} \cite{mokady2021clipcap} is a Transformer-based encoder-decoder model. It utilizes CLIP as an image encoder, and using a mapping network, maps image embeddings to a prefix which is used by a text-decoder, namely GPT2, to generate a caption. The pre-trained modules and the freshly-initialized mapping network are simply fine-tuned during the training process. 

From the set of baseline models described above, we fine-tuned ViT+RoBERTA, ViT+GPT2, BEiT+GPT2, GIT, BLIP and CLIPCap on the training set of our dataset. To understand zero-shot performance for figure-captioning task, we evaluated Pegasus, TrOCR, PromptCap and Flamingo-mini models by using their pretrained weights for inference without fine-tuning them on our dataset. 

For all fine-tuning experiments, we used AdamW  optimizer with $\beta_1 = 0.9$ \& $\beta_2 = 0.99$. We fine-tuned ViT+RoBERTA, ViT+GPT2, BEiT+GPT2 for 5 epochs with batch size 8. We used a linear rate scheduler with an initial learning rate of $2e-5$; generation was handled using a greedy strategy. For training GIT, BLIP and CLIPCap models, we used a learning rate of $1e-5$ and used nucleus sampling for text generation during inference.

\subsection{ Qualitative analysis}

In this section, we provide a detailed qualitative analysis of the output of BLIP-RLHF and BLIP (Fine-tuned) models.

\textbf{Comparative analysis:}
In the first example shown at the top left in Figure~\ref{fig:RLHF-vs-base}, we see that the generated caption with the base model BLIP has many issues.
For instance, it seems to have identified the word ``edges'' from the name of the model ``Deep-Edge'' used in the figure, despite that the figure does not actually show the number of edges in each experiment as the caption mentions. 
Instead, it shows the average epoch time in seconds for each of the different experiments, which is roughly captured by the BLIP-RLHF caption.
% In the second example shown at the top right of Figure~\ref{fig:RLHF-vs-base}, 
In the second example shown in the middle of Figure~\ref{fig:RLHF-vs-base}, 
the BLIP model completely hallucinates the caption whereas the BLIP-RLHF caption reveals the essence of the figure while also seemingly using the semantics of this specific chart-type, \eg, the phylogenetic tree shows the evolutionary relationships between different groups of fish and from the phylogenetic tree we can see how large each group is and the similarities between the groups of fish as well.
This also illustrates the ability of our approach to generalize to a variety of different chart types as we only obtained actual human feedback for line charts.
For the captions generated for the chart shown at the right in Figure~\ref{fig:RLHF-vs-base}, we see that BLIP generates a completely useless caption that has no alignment with the actual chart.
In comparison, the caption generated using BLIP-RLHF mentions the estimated and actual curves present in the chart while also correctly indicating that these curves are plotted in terms of time.
Most strikingly, the generated caption refers to the curves using their color (\ie, red line, blue dots), hence, the generated caption not only mentions important text from the chart, but also refers to the visual properties of the curves when mentioning them in the generated caption. 

\textbf{Human-Evaluation of model generated captions}:
To further evaluate the generated captions, we conducted a small-scale human evaluation experiment. Specifically, we randomly select 100 figures from the Test set of our proposed benchmark and generate captions using the BLIP and BLIP-RLHF models. We present the triplet of Figure, corresponding BLIP, and BLIP-RLHF generated captions (after randomizing the order of the two captions) to 10 human subjects. Each human subject is asked to rank the two captions based on which caption they think is better. We ask the subjects to specifically consider helpfulness, visual-descriptiveness, OCR alignment, and takeaway while ranking individual pairs of captions. To guide the subjects, we first explain each metric [helpfulness, visual-descriptiveness, OCR alignment, and takeaway] and present each human subject with 100 samples from our human-annotated dataset with individual figures, ground truth caption, and the corresponding metric scores (recorded in 5-point Likert scale). From our study, we find that on average 85\% of the time, BLIP-RLHF generated caption was selected as the better caption relative to BLIP generated caption. From our small-scale study, we conclude that RLHF does improve the quality of the captions when compared to fine-tuning existing Vision Language models for the task of figure-caption generation.

\section{Datasheet}

\subsection{Motivation}
\textbf{For what purpose was the dataset created?} We created this dataset to provide researchers ability to develop and evaluate their respective figure-to-caption generation pipelines for reader preference aligned caption generation.

\textbf{Who created the dataset (e.g., which team, research group) and on behalf of which entity(e.g., company, institution, organization)?} We would provide the details of the authors upon acceptance of the paper, due to double-blind review process.

\textbf{Who funded the creation of the dataset?} No funding was recieved in any form in creation of this dataset.

\subsubsection{Author Statement}
The authors of this paper bear all responsibilities for the distribution, and maintenance of our proposed dataset. This document follows the Datasheet format \cite{gebru2021datasheets} whenever applicable.

\subsection{Distribution}

\textbf{Will the dataset be distributed to third parties outside of the entity (e.g., company, institution, organization) on behalf of which the dataset was created?} Yes, the dataset is public and available for usage on the internet.

\textbf{How will the dataset will be distributed (e.g., tarball on website, API, GitHub)?} The dataset and the corresponding codebase used in generating the dataset is avaialable through following links:\\
\small
\textbf{Benchmark:} \url{https://doi.org/10.6084/m9.figshare.23504517}\\
\small
\textbf{Code:} \url{https://github.com/FigCapsHF/FigCapsHF}\\
\small
\textbf{Documentation:} \url{https://figcapshf.github.io/}\\

\textbf{Have any third parties imposed IP-based or other restrictions on the data associated with the
instances?} No.

\textbf{Do any export controls or other regulatory restrictions apply to the dataset or to individual instances?} No.

\subsection{Maintenance}

\textbf{Who will be supporting/hosting/maintaining the dataset?} The authors will be supporting, hosting and maintaining the dataset.

\textbf{How can the owner/curator/manager of the dataset be contacted (e.g., email address)?} We would provide the details of the contact persons upon acceptance of the paper, due to double-blind review process.

\textbf{Is there an erratum?} No. We will accordingly make announcements if there is any.

\textbf{Will the dataset be updated (e.g., to correct labeling errors, add new instances, delete instances)?} Yes. Announcements regarding any updates to dataset and code would be posted here: \url{https://github.com/FigCapsHF/FigCapsHF}

\textbf{If the dataset relates to people, are there applicable limits on the retention of the data associated with the instances (e.g., were the individuals in question told that their data would be retained for a fixed period of time and then deleted)?} N/A

\textbf{Will older versions of the dataset continue to be supported/hosted/maintained?} Yes.

\textbf{If others want to extend/augment/build on/contribute to the dataset, is there a mechanism for them to do so?} Yes. 

\subsection{Composition}

\textbf{What do the instances that comprise the dataset represent?} Please refer to section \ref{format} for detailed description of the dataset composition.

\textbf{How many instances are there in total (of each type, if appropriate)?} in total we have 06,834 training pairs, 13,354 validation pairs, and 13,355 test figure-caption pairs with feedback scores.

\textbf{Does the dataset contain all possible instances or is it a sample (not necessarily random) of
instances from a larger set?} The dataset contain all possible instances

\textbf{Is there a label or target associated with each instance?} Yes. Each figure image in the dataset has a corresponding caption and a set of values representing the predicted feedback score for metrics \textbf{('helpfulness', 'ocr', 'visual', 'takeaway'}.

\textbf{Is any information missing from individual instances?} No.

\textbf{Are relationships between individual instances made explicit (e.g., users’ movie ratings, social network links)?}  N/A

\textbf{Are there recommended data splits (e.g., training, development/validation, testing)?} Yes. The dataset consists of 3 splits: Train, Validation and Test. We have explicitly provided individual splits as separate data folders.

\textbf{Are there any errors, sources of noise, or redundancies in the dataset?} No.

\textbf{Is the dataset self-contained, or does it link to or otherwise rely on external resources (e.g., websites, tweets, other datasets)?} The dataset is entirely self-contained and does not require any external resource.

\textbf{Does the dataset contain data that might be considered confidential?} No.

\textbf{Does the dataset contain data that, if viewed directly, might be offensive, insulting, threatening,or might otherwise cause anxiety?} No.

\subsection{Collection Process}

\textbf{Who was involved in the data collection process (e.g., students, crowdworkers, contractors) and
how were they compensated (e.g., how much were crowdworkers paid)?} The authors were involved in the curation of the data obtained from a publicaly avaialbe source.

\textbf{Over what timeframe was the data collected?} Februray 2023-May 2023

\subsection{Uses}

\textbf{Has the dataset been used for any tasks already?} Our work on human feedback aligned figure caption generation uses the proposed dataset. 

\textbf{Is there a repository that links to any or all papers or systems that use the dataset?}  N/A

\textbf{What (other) tasks could the dataset be used for?} Evaluating image-to-text generation models for a domain specific performance. 

\textbf{Is there anything about the composition of the dataset or the way it was collected and
preprocessed/cleaned/labeled that might impact future uses?} No.

\subsection{Data Format} \label{format}
\begin{figure}[t]
\includegraphics[width=\linewidth]{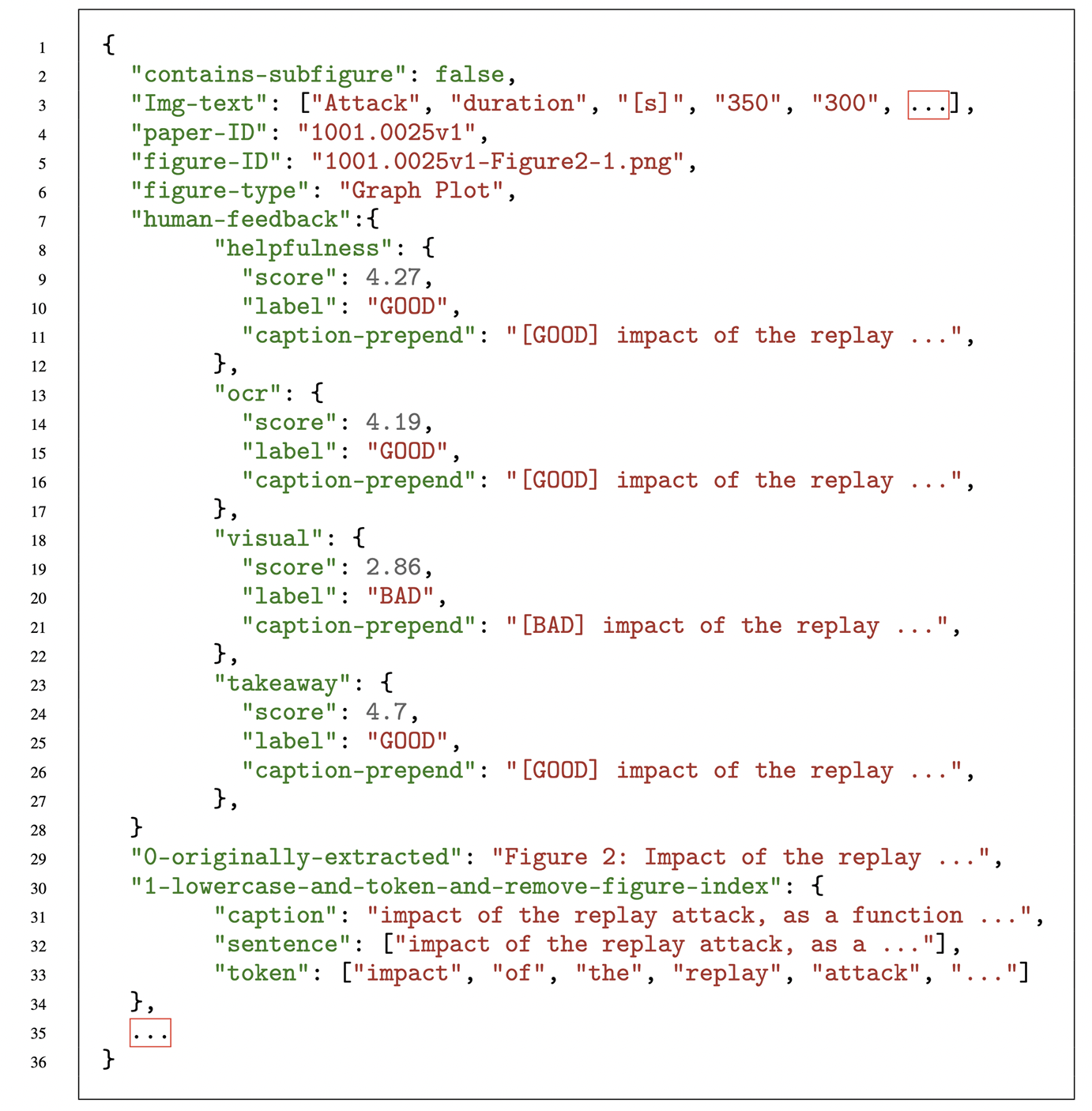}
\caption{ Human Feedback Benchmark Data Example for Figure-Caption Generation with RLHF 
}
\label{benchmark-HF-data-example}
\end{figure}

% TODO: The dataset itself should ideally use an open and widely used data format. Provide a detailed explanation on how the dataset can be read. For simulation environments, use existing frameworks or explain how they can be used.
For each figure-caption pair, the figure-image is stored as a PNG, and the figure-caption (with associated metadata) is stored in a JSON format. ~\ref{benchmark-HF-data-example} is an example from the dataset.

In each figure-caption's metadata file, the fields are: 

\begin{itemize}

\item
  \textbf{contains-subfigure:} boolean (if figure-image contains
  subfigures)
\item
  \textbf{paper-ID:} the unique paper ID in the arXiv dataset
\item
  \textbf{figure-ID:} the extracted figure ID of paper (the index is not
  the same as the label in the caption)
\item
  \textbf{figure-type:} the figure type
\item
  \textbf{0-originally-extracted:} original figure-caption
  \begin{itemize}
\item
  \textbf{caption:} caption after each normalization
\item
  \textbf{sentence:} a list of segmented sentences
\item
  \textbf{token:} a list of tokenized words
\end{itemize}
\item
  \textbf{1-lowercase-and-token-and-remove-figure-index:} Removed figure
  index and the captions in lowercase
  \begin{itemize}
      \item Same substructure as 0-originally-extracted
  \end{itemize}
  
\item
  \textbf{2-normalized:}

  \begin{itemize}
  \item
    \textbf{2-1-basic-num:} caption after replacing the number
     \begin{itemize}
      \item Same substructure as 0-originally-extracted
  \end{itemize}
  \item
    \textbf{2-2-advanced-euqation-bracket:} caption after replacing the
    equations and contents in the bracket
     \begin{itemize}
      \item Same substructure as 0-originally-extracted
  \end{itemize}
  \end{itemize}
\item
  \textbf{Img-text:} texts extracted from the figure, such as the texts
  for labels, legends \ldots{} etc.
\end{itemize}

Within the "human-feedback" field, we have the inferred human-feedback for the different metrics (helpfulness, ocr, takeaway, and visual). The tokens are decided based on the median score of the dataset on that metric.

\begin{itemize}
    \item \textbf{Helpfulness:} Expert's rating on how helpful a caption is to understand a scientific figure
    \begin{itemize}
        \item \textbf{Score:} predicted score
        \item \textbf{Token:} {[}Good{]}/{[}Bad{]} 
        \item \textbf{caption-prepend:} 1-lowercase-and-token-and-remove-figure-index caption with the token
    \end{itemize}
    \item \textbf{Takeaway:} Expert's rating on the takeaway from the scientific image
    \begin{itemize}
        \item Same substructure as Helpfulness
    \end{itemize}
    \item \textbf{OCR:} Expert's rating on the OCRs expressiveness
    \begin{itemize}
        \item Same substructure as Helpfulness
    \end{itemize}
    \item \textbf{Visual:} Expert's rating on the visualness of the
scientific figure
    \begin{itemize}
        \item Same substructure as Helpfulness
    \end{itemize}
\end{itemize}

\subsubsection{Reading Data}
% TODO: The dataset itself should ideally use an open and widely used data format. Provide a detailed explanation on how the dataset can be read. 
For all figure-caption pairs, all of the figure-images are in their respective train/val/test subfolders under the "No-Subfig-Img" folder. The corresponding figure-captions and associated metadata are in their respective train/val/test subfolders under the "Caption-All' folder, bearing the same filename as their image. In order to read the data, one can read the file-names of all the figure-images in a particular data-split, and retrieve the corresponding figure-caption metadata using the image file-names (instead iterating through the captions also works). Another approach is to iterate through the "file\_idx.json" file under the "List-of-Files-for-Each-Experiments/First-Sentence/(train/val/test)"  folder, which contains a list of all image-names we used for that data split. 

\subsubsection{Reproducibility}
% TODO: For benchmarks, the supplementary materials must ensure that all results are easily reproducible. Where possible, use a reproducibility framework such as the ML reproducibility checklist, or otherwise guarantee that all results can be easily reproduced, i.e. all necessary datasets, code, and evaluation procedures must be accessible and documented.

We have provided easy access to the benchmark dataset which was used to conduct all of our experiments, including the augmented caption that was used during RLHF fine-tuning. 

We have also provided access to a github repository, which contains the code used to: train a baseline,  fine-tune a model using human-feedback, and evaluate the model on the test set.

\end{document}